\documentclass{article}

\usepackage[preprint]{neurips_2026}
\usepackage{amssymb}
\usepackage{multirow}
\usepackage[utf8]{inputenc} %
\usepackage[T1]{fontenc}    %
\usepackage{hyperref}       %
\usepackage{url}            %
\usepackage{booktabs}       %
\usepackage{amsfonts}       %
\usepackage{nicefrac}       %
\usepackage{microtype}      %
\usepackage{xcolor}         %
\usepackage{amsmath}   %
\usepackage{amsfonts}  %
\usepackage{graphicx}
\usepackage{amsthm}
\usepackage{algorithmicx,algpseudocode}
\usepackage[subfigure]{tocloft}
\usepackage{subfigure}
\usepackage{algorithm,algpseudocode}
\usepackage{empheq}
\usepackage{titletoc}
\usepackage{float}
\theoremstyle{remark}

\newcommand{\Rho}{\mathrm{P}}

\DeclareMathOperator*{\argmin}{arg\,min}
\title{Neural Galerkin Normalizing Flows for Bayesian Inference of Diffusions with Inaccessible Boundaries}

\author{%
  Riccardo Saporiti \\
  CSQI\\
  \'Ecole Polytechnique F\'ed\'erale de Lausanne\\
  Lausanne, Switzerland 1015 \\
  \texttt{riccardo.saporiti@epfl.ch} \\
  \And
  Fabio Nobile \\
  CSQI\\
  \'Ecole Polytechnique F\'ed\'erale de Lausanne\\
  Lausanne, Switzerland 1015 \\
  \texttt{fabio.nobile@epfl.ch} \\
}

\begin{document}

\maketitle

\begin{abstract}
One of the primary challenges in Bayesian inference on the parameters of a diffusion model from discrete observations is the unavailability of an analytical expression for the transition density function between consecutive observation times, which is needed to derive the likelihood function.

Extending previous studies that solve Fokker-Planck (FP) type partial differential equations with Normalizing Flows, we propose a new Normalizing Flow architecture to learn the transition density function of the diffusion process between two observation times. We do so by solving in a Neural Galerkin framework the associated FP equation with a Dirac mass as initial condition, over a specified training distribution of the 
initial datum and the coefficients of the diffusion. We specifically focus on processes whose diffusion matrix vanishes in certain inaccessible boundary regions, such as Stochastic Volatility models that satisfy a Feller condition.

The product of the obtained transition densities evaluated along the observed trajectory approximates the likelihood function, thereby enabling cheap posterior sampling via Markov chain Monte Carlo (MCMC). After the offline training phase, inference becomes significantly more efficient, as it avoids the need to solve the FP equation in real time for each parameter proposed by the MCMC sampler or to rely on other likelihood-free methods for Bayesian inference that involve repeated simulation of diffusion bridges.

\end{abstract}

\section{Introduction}

One of the most challenging aspects of performing Bayesian Inference on the parameters of discretely observed diffusion processes is the unavailability of an analytic expression of the likelihood function.
This limitation has spurred research to develop positive estimators for the likelihood function through numerical integration of the diffusion process or exact simulation techniques \cite{beskos2006exact, https://doi.org/10.1111/j.1467-9469.2012.00812.x}.
However, exact simulation of diffusion processes between two observed values is notoriously difficult, especially for nonlinear, multivariate Stochastic Differential Equations (SDEs). 
In many cases, it is necessary to use numerical discretization by adding imputed points between two observation time instants \cite{eraker2001mcmc,elerian2001likelihood}. 
Depending on the focus of the inference, either Metropolis-within-Gibbs, Pseudo-Marginal MCMC, or Particle MCMC algorithms are usually employed \cite{508a6095-c33f-3d5d-9a9a-f004c0c0a59a, AndrieuDoucetHolenstein2010, Fuchs}. These approaches quickly become very costly, particularly in low-frequency observation regimes, where many imputed points are needed, potentially leading to unwanted correlations in the chain \cite{roberts2001inference}.

A less explored direction is to seek an explicit approximation of the Transition Probability Density Function (TPDF) between two observations. Such a TPDF is the solution of the well-known Fokker-Planck (FP) partial differential equation (PDE) \cite{Gikhman_Skorokhod_1972}. This equation poses significant numerical challenges due to positivity and integrability constraints, and because its initial condition is a Dirac delta distribution, which is difficult to approximate numerically.

Since the likelihood of a trajectory composed of $n+1$ discrete observations factors into the product of $n$ TPDFs, the computational cost for a single likelihood evaluation grows with the length of the time series. This computational burden must be incurred for each sample generated by an MCMC-type algorithm that walks towards the posterior distribution of the SDE parameters given the data. Therefore, developing efficient numerical algorithms for fast likelihood evaluation is vital for several real-time applications.

Nowadays, Deep Learning algorithms have enabled the solution of (parametric) PDEs that are intractable for traditional discretization methods \cite{RAISSI2019686, SIRIGNANO20181339,doi:10.1073/pnas.1718942115, karniadakis2021piml, Quarteroni_2025}.
However, the use of Deep Learning for transition density estimation is still in its early stages. In \cite{su2024deep}, Physics-Informed Neural Networks (PINNs) are used to compute parametric approximations of the Kolmogorov Backward equation that solves for the cumulative density function. This method does not ensure that the numerical solution represents a valid density function, which could limit its applicability in Bayesian inference for diffusion processes.

Normalizing Flows (NFs) have demonstrated promising results in solving FP equations within data-driven frameworks \cite{lu2022learning} and alongside PINN-based approaches \cite{TANG2022111080, Xiaodong_Feng2022-ak}.
Nonetheless, PINNs, which train a neural network uniformly in the space-time domain, struggle to enforce the causality of the numerical solution and are notoriously prone to failure in advection-dominated regimes \cite{krishnapriyan2021characterizingpossiblefailuremodes, pmlr-v202-daw23a}. In this context, concentrated initial conditions further complicate the issue by contributing to the emergence of highly spatially localized features in the solution. 

To approximate TPDFs associated with FP equations with no boundaries,  the work \cite{saporiti2026neural} used NFs combined with the Neural Galerkin framework of \cite{BRUNA2024112588}, yielding Neural Galerkin Normalizing Flows (NGNFs). The authors proposed using real-valued non-volume preserving transformations to approximate the TPDF parametrically with respect to the position of the Dirac delta initial condition. 
 In contrast to traditional PINNs, NGNFs avoid solving the PDE over the entire space-time domain and, through an adaptive sampling strategy, propagate the NF's parameters forward in time, naturally preserving causality. 

In this work, we follow a similar path but consider a different NF architecture that can accommodate inaccessible boundaries. 
Moreover, we propose a Bayesian inference methodology based on NGNF to infer the parameters of the SDE, and validate it on Stochastic Volatility (SV) models, a class of non-uniformly elliptic diffusion processes linked to an advection-dominated FP equation that poses significant challenges for traditional numerical solvers.

\subsection{Existing methodologies for Bayesian inference of diffusion processes}

Since the TPDF is generally unavailable, the literature has proposed several alternatives for Bayesian Inference, including a) discretization-based MCMC \cite{roberts2001inference, 10.1111/rssb.12497,Golightly2022,jasra2025bayesianinferencenonsynchronouslyobserved}, b) exact simulation \cite{beskos2006exact, https://doi.org/10.1111/j.1467-9469.2012.00812.x,sant2025bayesian, garcia2022unbiased}, c) likelihood-free methods \cite{Picchini02102014, 10.1214/24-BA1467}, d) variational inference \cite{ryder2018black} and e) closed-form expansions \cite{ait_sahalia2008closed, choi2015explicit, YANG2019256, iguchi2025closed}.
While these approaches can be effective for specific applications, they also present certain drawbacks. These may include high online computational costs (a, b, c), bias associated with discretization of the SDE (a, c, d), or limited applicability (b, e), underscoring the need for scalable likelihood surrogates. 

\subsection{Contributions}

\begin{itemize}
    \item We build on the NGNF framework developed in \cite{saporiti2026neural} and introduce a new class of bounded Normalizing Flows to solve parametric FP equations with vanishing diffusion at inaccessible boundaries. This approach effectively addresses advection-dominated PDEs. The NF is parameterized by the SDE parameters and the position of the initial Dirac mass to enable statistical inference.
    \item 
    Unlike purely data-driven methods, 
    such as those based on learning the NF from forward simulation of the SDE, 
    our approach more strongly enforces the structural properties of a flow of probability measures and, thanks to adaptive sampling schemes, allows for coherent modeling of the relationships among the diffusion parameters, the initial condition, and the final solution.
    \item We propose a Bayesian inference procedure based on our NGNFs and evaluate its performance on fully observed SV models. 
   Our results show that NFs enable fast likelihood evaluation and robustness in low-frequency observation regimes, thereby considerably accelerating Bayesian inference for diffusion models.
\end{itemize}

\section{Problem Formulation}\label{Section_contribution}
Let $\boldsymbol{X}(t)$ be a $d$-dimensional stochastic process satisfying an SDE of the form
\begin{equation} \label{eq:ngnf_inf_Diffusion_process_mu}
    d\boldsymbol{X}(t) = \boldsymbol{b}(t,\boldsymbol{X}(t),\boldsymbol{\mu})dt + \sqrt{\Sigma(t,\boldsymbol{X}(t),\boldsymbol{\mu})}d\boldsymbol{W}(t) \quad \text{for $t\in(0,T]$},
\end{equation}
where $T>0$, $\boldsymbol{b}:[0,T]\times\Omega\times\Xi\rightarrow\mathbb{R}^{d}$ is a drift function, $\Sigma:[0,T]\times\Omega\times\Xi\rightarrow\mathbb{R}^{d\times d}$ is a semi-positive definite matrix, $\boldsymbol{\mu}\in\Xi\subset \mathbb{R}^{p}$ is a $p$ dimensional vector of parameters, $\boldsymbol{W}(t)$ is a $d$-dimensional vector of uncorrelated Brownian Motions and $\Omega\subseteq\mathbb{R}^d$ is the domain of $\boldsymbol{X}(t)$. We assume that the domain of the diffusion process is of the form:
\begin{equation}\label{eq:ngnf_inf_domain}
    \Omega = \{ \boldsymbol{x} \in \mathbb{R}^d : a_m < x_m \quad \text{for } m = 1, \ldots, d \},
\end{equation}
where \(a_m \in \mathbb{R} \cup \{-\infty\}\), and 
the diffusion matrix $\Sigma$ degenerates on $\partial\Omega$. This condition is typical of SV models, where the volatility of the asset is the state variable that drives the determinant of the diffusion matrix to zero.

Let $T>t>s\geq0$. The transition probability density function (TPDF) of $\boldsymbol{X}(t)|\boldsymbol{X}(s)=\boldsymbol{x_0}\in\Omega$ is denoted by $\rho(\boldsymbol{x}|t,s,\boldsymbol{x_0})$
and satisfies $\mathbb{P}(\boldsymbol{X}(t)\in A|\boldsymbol{X}(s)=\boldsymbol{x_0})=\int_{A}\rho(\boldsymbol{x}|t,s,\boldsymbol{x_0})d\boldsymbol{x}$ for any Borel set $A\subset\Omega$.
It is well-known that the TPDF
evolves according to the Fokker-Planck equation 
\cite{Gikhman_Skorokhod_1972}
\begin{subequations}\label{eq:ngnf_inf_TPDF_Fokker_Planck_mu}
\begin{empheq}[left=\empheqlbrace]{align}
\label{eq:ngnf_inf_TPDF_Fokker_Planck_a}
\partial_{t}\rho(\boldsymbol{x}|t,s,\boldsymbol{x_0},\boldsymbol{\mu})
&= \mathcal{L}^{\star}_t\left(\rho(\cdot|t,s,\boldsymbol{x_0},\boldsymbol{\mu})\right)(\boldsymbol{x})
&& \text{for $\boldsymbol{x}\in\Omega,\: t\in(s,T]$}, \\
\label{eq:ngnf_inf_TPDF_Fokker_Planck_b}
\rho(\boldsymbol{x}|s,s,\boldsymbol{x_0},\boldsymbol{\mu})
&= \delta_{{x}_0}(\boldsymbol{x})
&& \text{for $\boldsymbol{x}\in\Omega$}, \\
\label{eq:ngnf_inf_TPDF_Fokker_Planck_c}
\int_{\Omega} \rho(\boldsymbol{x}|t,s,\boldsymbol{x_0},\boldsymbol{\mu})\,d\boldsymbol{x}
&= 1
&& \text{for $t\in(s,T]$}, \\
\label{eq:ngnf_inf_TPDF_Fokker_Planck_d}
\rho(\boldsymbol{x}|t,s,\boldsymbol{x_0},\boldsymbol{\mu})
&\geq 0
&& \text{for $\boldsymbol{x}\in\Omega,\:t\in(s,T]$},
\end{empheq}
\end{subequations}
where $\delta_{{y}}( \cdot )$ denotes the Dirac delta distribution centered at $\boldsymbol{y}\in\mathbb{R}^{d}$ and $\mathcal{L}^{\star}$ is the $L^{2}$ adjoint of the generator of the Markov Process $\boldsymbol{X}(t)$, which reads, for $f\in\mathcal{C}^2(\Omega;\mathbb{R})$
\begin{equation} \label{eq:Adjoint_generator_mu}
    \mathcal{L}^{\star}_t(f(\cdot))(\boldsymbol{x}) = \nabla_{\boldsymbol{x}}\cdot [ -\boldsymbol{b}(t,\boldsymbol{x},\boldsymbol{\mu})f({\boldsymbol{x}})+\frac{1}{2}\nabla_{\boldsymbol{x}}\cdot(\Sigma(t,\boldsymbol{x},\boldsymbol{\mu})f(\boldsymbol{x})) ],
\end{equation}
see e.g. \cite{Stochastic_Processes_and_Applications}.
In this work, we focus on the case where the boundary \(\partial\Omega\) is inaccessible, and we specifically treat SV models that satisfy a strict Feller condition. Under these assumptions, the solution $\rho(\cdot|t,s,\boldsymbol{x_0},\boldsymbol{\mu})$ is identically zero on $\partial\Omega$, leading to a boundary condition of homogeneous Dirichlet type \cite{Lucic01102012, doi:10.1142/S2424786320500486}.

\section{Nonlinear parametrization of the Transition Probability Density Function} \label{sect:ngnf_inf_Section_nonlinear_parametrization}
Normalizing Flows are diffeomorphisms that transform the probability density function (PDF) of a source distribution into that of a target distribution, often using parameterizations based on neural networks. 
Following the Neural Galerkin framework, we represent the approximated solution of the FP PDE at a specific time $t$ by the time-dependent vector of parameters $\boldsymbol{\theta}(\tau)\in\Theta\subseteq\mathbb{R}^{M}$ that characterizes the architecture of the employed Normalizing Flow, namely
\begin{equation} \label{eq:ngnf_inf_pre_parametrization_TPDF}
\rho(\boldsymbol{x}|t,s,\boldsymbol{x_0},\boldsymbol{\mu})\simeq\Rho(\boldsymbol{x}|\boldsymbol{\theta}(\tau),s,\boldsymbol{x_0},\boldsymbol{\mu}),
\end{equation}
where $\tau=t-s$, and $\Rho$ represents the PDF (parametric in $s,\boldsymbol{x_0},\boldsymbol{\mu}$) generated by the Normalizing Flow with parameter vector $\boldsymbol{\theta}(\tau)$. For further details, see section \ref{sect:ngnf_inf_section_nf_structure}.

We assume that the observations of the stochastic process \eqref{eq:ngnf_inf_Diffusion_process_mu} are collected within the time interval \([0, T]\). For a fixed $s\in[0,T-\Delta]$, we solve the FP PDE over the time horizon $[s,s+\Delta]$, where $\Delta > 0$ represents the maximum time between observations. In turn, this implies that $\tau\in[0,\Delta]$.
For convenience, we incorporate $s$ into the vector $\boldsymbol{\mu}$, defining 
$\boldsymbol{\tilde{\mu}}=[\boldsymbol{\mu},s]\in\Xi^{\star}=\Xi\times[0,T-\Delta]\subseteq \mathbb{R}^{p+1}$. Hence, the NF's PDF  \eqref{eq:ngnf_inf_pre_parametrization_TPDF} is rewritten as $\Rho(\boldsymbol{x}|\boldsymbol{\theta}(\tau),\boldsymbol{x_0},\boldsymbol{\tilde{\mu}})$.

\subsection{System of ODEs for Normalizing Flow's parameters}
\label{parameter_update_equation_derivation}
 
Following the approach outlined in \cite{BRUNA2024112588, saporiti2026neural}, we extend the derivation of the system of ODEs describing the time evolution of the parameters $\boldsymbol{\theta}(\tau)$ to the $\boldsymbol{\tilde{\mu}}$-parametric setting.  We substitute the ansatz \eqref{eq:ngnf_inf_pre_parametrization_TPDF} into the PDE for the TPDF \eqref{eq:ngnf_inf_TPDF_Fokker_Planck_mu}. Assuming that $\boldsymbol{\theta}(\tau)$ is differentiable, we obtain the residual function $r_{t,s}:\Theta\times\dot{\Theta}\times\Omega\times\Omega\times\Xi^{\star}\rightarrow\mathbb{R}$ defined as
\begin{equation} \label{eq:ngnf_inf_residual_Neural_Galerkin}
    r_{t,s}(\boldsymbol{\theta},\boldsymbol{\zeta},\boldsymbol{x},\boldsymbol{x_0},\boldsymbol{\tilde{\mu}}) = \nabla_{\boldsymbol{\theta}}\Rho(\boldsymbol{x}|\boldsymbol{\theta}(\tau),\boldsymbol{x_0},\boldsymbol{\tilde{\mu}})\cdot \boldsymbol{\zeta} - \mathcal{L}^{\star}_t(\Rho(\cdot|\boldsymbol{\theta}(\tau),\boldsymbol{x_0},\boldsymbol{\tilde{\mu}}))(\boldsymbol{x}),
\end{equation}
where $\dot{\Theta}$ is the set of time derivatives of $\boldsymbol{\theta}$.
The evolution of the parameters of the Normalizing Flow is given by the following optimization problem, which has to be solved at each time instant:
\begin{equation} \label{eq:ngnf_inf_optimization_Neural_Galerkin}
    {\boldsymbol{\dot{\theta}}}(\tau) \in \argmin_{\boldsymbol{\zeta}\in\dot{\Theta}} J_{\tau}(\boldsymbol{\theta},\boldsymbol{\zeta}), 
\end{equation}
where the objective $J_{\tau}:\Theta\times\dot{\Theta}\rightarrow\mathbb{R}$ is defined as
\begin{equation} \label{eq:ngnf_inf_J_cost_functional_Neural_Galerkin}
    J_{\tau}(\boldsymbol{\theta},\boldsymbol{\zeta})=\int_{\Xi^{\star}}\int_{\Omega}\int_{\Omega} \lvert r_{s+\tau,s}(\boldsymbol{\theta},\boldsymbol{\zeta},\boldsymbol{x},\boldsymbol{x_0},\boldsymbol{\tilde{\mu}}) \rvert ^{2} d\nu_{{\theta}(\tau)}(\boldsymbol{x}|\boldsymbol{x_0},\boldsymbol{\tilde{\mu}}) d\eta(\boldsymbol{x_0})d\tilde{\pi}_0^{\star}(\boldsymbol{\tilde{\mu}}). 
\end{equation}

In \eqref{eq:ngnf_inf_J_cost_functional_Neural_Galerkin}, $\eta$ and $\nu_{{\theta}(\tau)}$ are positive measures over $\hat{\Omega} \subseteq \Omega$, where $\hat{\Omega}$ is the support of the PDFs generated by the NF, which will be specified later.
$\tilde{\pi}_0^{\star}$ denotes the joint distribution that tensorizes a distribution over the parameters of the SDE  \eqref{eq:ngnf_inf_Diffusion_process_mu}, denoted as $\tilde{\pi}_0$, with the uniform distribution $\mathcal{U}(\cdot;[0,T-\Delta])$. 
Thanks to the properties of the map defined by the Normalizing Flow at time $t$, we can sample progressively from the evolving approximation $\Rho$, which allows us to take $\nu_{{\theta}(\tau)}(\cdot|\cdot,\cdot)=\Rho(\cdot|\boldsymbol{\theta}(\tau),\cdot,\cdot)$. Similar choices are employed in \cite{BRUNA2024112588, saporiti2026neural} and are crucial for tackling the curse of dimensionality in advection-dominated PDEs. 
Numerically, \eqref{eq:ngnf_inf_J_cost_functional_Neural_Galerkin} is approximated by Monte Carlo sampling, resulting in a least squares problem that is solved by using LSMR \cite{FongSaunders2011}.

\section{Normalizing Flow for Transition Density Estimation}
\label{sect:ngnf_inf_section_nf_structure}

\subsection{Background on Normalizing Flows}\label{sect:ngnf_inf_section_background}

  NFs are bijective, differentiable transformations $\boldsymbol{n}_{\theta}:\mathcal{X}\rightarrow \mathcal{Z}$ used to relate the PDF of a target random variable $\boldsymbol{X}\in\mathcal{X}\subseteq\mathbb{R}^d$ to that of a source (reference) random variable $\boldsymbol{Z}\in\mathcal{Z}\subseteq\mathbb{R}^d$. 
For a given set of parameters $\boldsymbol{\theta}\in\Theta\subseteq{\mathbb{R}^{M}}$, the density of the target random variable is expressed through the change of variables formula 
\begin{equation} \label{eq:ngnf_inf_NF_identity}
    p_{X}(\boldsymbol{x})= p_{Z}(\boldsymbol{n}_{\theta}(\boldsymbol{x}))\lvert \mathrm{det}(\nabla_{\boldsymbol{x}}\boldsymbol{n}_{\theta}(\boldsymbol{x})) \rvert,
\end{equation}
where $p_{X}$ and $p_Z$ are the PDFs of $\boldsymbol{X}$ and $\boldsymbol{Z}$ respectively. 

According to this formula, we can obtain samples from $\boldsymbol{X}$ by mapping back samples from the reference distribution $\boldsymbol{z}\sim \boldsymbol{Z}$ through the Flow, such that $\boldsymbol{x}=\boldsymbol{n}_{\boldsymbol{\theta}}^{-1}(\boldsymbol{z})$. 

In this work, we utilize the identity given by the change of variable formula \eqref{eq:ngnf_inf_NF_identity} to represent the solution of the FP equation. This immediately ensures that the numerical solution satisfies the properties described in \eqref{eq:ngnf_inf_TPDF_Fokker_Planck_c}, \eqref{eq:ngnf_inf_TPDF_Fokker_Planck_d}. 

We work under the following assumption: the target TPDF decays at infinity more rapidly than any polynomial. This assumption is reasonable for SDEs with coefficients that grow at most linearly at infinity and allows us to approximate the TPDF using a Normalizing Flow with compact support. Therefore, we confine the TPDF realized by the NF to the hyperrectangle $ 
\hat{\Omega}=
\bigotimes_{m=1}^d (\hat{a}_m,\hat{b}_m)
$, where $\boldsymbol{\hat{a}},\boldsymbol{\hat{b}}\in\mathbb{R}^d$ and 
$-\hat{a}_m$, $\hat{b}_m$ are large enough for all $m\in\{1,...,d\}$ except those for which $a_m>-\infty$ (cf. equation \eqref{eq:ngnf_inf_domain}), where we set $\hat{a}_m=a_m$.
With this construction, $\partial\hat{\Omega}$ contains a portion of the true boundary of the state space, namely
$\Gamma=\partial\Omega\cap\partial\hat{\Omega}\neq\emptyset$.
We solve the FP PDE on \(\hat{\Omega}\) by imposing homogeneous Dirichlet boundary conditions on \(\Gamma\) and zero-flux boundary conditions, corresponding to reflected diffusion,  on the artificial boundary \(\partial\hat{\Omega}\setminus\Gamma\).

\subsection{Truncated-Gaussian Normalizing Flows for probability density estimation} \label{Truncated_gaussian_NF}
 
For a fixed $\tau>0$, the source distribution $p_Z$ is transformed through a time-dependent mapping, defining an approximation to the TPDF that solves \eqref{eq:ngnf_inf_TPDF_Fokker_Planck_mu} as 
\begin{equation} \label{eq:ngnf_inf_FP_approximation_by_NF}
    \Rho(\boldsymbol{x}|\boldsymbol{\theta}(\tau),\boldsymbol{x_0},\boldsymbol{\tilde{\mu}})=p_{Z}(\boldsymbol{n}_{\theta(\tau)}(\boldsymbol{x}|\boldsymbol{x_0},\boldsymbol{\tilde{\mu}}))\lvert \mathrm{det}(\nabla_{\boldsymbol{x}}\boldsymbol{n}_{\theta(\tau)}(\boldsymbol{x}|\boldsymbol{x_0},\boldsymbol{\tilde{\mu}}))|.
\end{equation}

Our architecture models the mapping $\boldsymbol{n}_{\theta(\tau)}$ by stacking several bijections in sequence. The final transformation builds upon the composition of an integer number of layers, $L\geq1$, such that, for $m=0,...,d-1$  
\begin{equation} \label{eq:ngnf_inf_NF_mapping}
    \begin{cases}
        (\boldsymbol{n}_{\theta(\tau)}(\boldsymbol{x}|\boldsymbol{x_0},\boldsymbol{\tilde{\mu}}))_{m+1}={x}^{L}_{m+1} \\
        x^{l}_{m+1}={n}_{\theta^{l}(\tau)}^{l,m+1}({x}^{l-1}_{m+1}|\boldsymbol{x}_{1:m}^1,\boldsymbol{x_0},\boldsymbol{\tilde{\mu}})) \quad \text{for $l=1,...,L$} \\
        x^0_{m+1}=x_{m+1},
    \end{cases}
\end{equation}
where $\boldsymbol{x}_{1:0}^1=\emptyset$. 
The smooth, bounded Normalizing Flow is characterized by a nonlinear invertible mapping which, after rescaling $\overline{\hat{\Omega}}$ to $[-1,1]^{d}$, we recursively apply for $l=2,...,L$ and for $m=0,...,d-1$. We define the mapping using the cumulative density function (CDF) of a mixture of truncated Gaussians (MTG) random variables
\begin{equation} \label{eq:ngnf_inf_base_transformation}
    \tilde{F}:[-1,1]\rightarrow[-1,1],\quad \tilde{F}(x)= 2[\sum_{k=1}^{G} \alpha_{k}\Phi_{[-1,1]}(x|\mathrm{m}_{k},\sigma_{k})]-1, 
\end{equation}
where $\Phi_{[x,y]}(\cdot|\mathrm{m},\sigma)$ and $\phi_{[x,y]}(\cdot|\mathrm{m},\sigma)$ denote the CDF and PDF, respectively, of a Gaussian random variable truncated to the interval $[x,y]$, with $\mathrm{m}$ and $\sigma$ representing the mean and standard deviation of the parent Normal distribution \cite{johnson1995continuous}. 
Transformation \eqref{eq:ngnf_inf_NF_mapping} involves an affine rescaling followed by $L-1$ applications of \eqref{eq:ngnf_inf_base_transformation} to a one-dimensional input vector while conditioning on the other state variables. 
 The first layer maps the NF's support to the unit-hypercube $[-1,1]^{d}$.
In formulas, for $m=0,...,d-1$,
\begin{equation} \label{eq:ngnf_inf_rescaling_1_1}
{n}_{\theta^1(\tau)}^{1,m+1}({x}_{m+1}|\boldsymbol{x}_{1:m}^1,\boldsymbol{x_0},\boldsymbol{\mu})=2\frac{{x}_{m+1}-\hat{a}_{m+1}}{\hat{b}_{m+1}-\hat{a}_{m+1}}-1.
\end{equation}
The conditioning variable $\boldsymbol{x_0}$ is also processed by \eqref{eq:ngnf_inf_rescaling_1_1}.

The dependency of the mapping on the parameters of \eqref{eq:ngnf_inf_Diffusion_process_mu}, the location of the initial condition \eqref{eq:ngnf_inf_TPDF_Fokker_Planck_mu}, and the observation time $s$ is  introduced through $d(L-1)$ Neural Networks; one for every layer $l=2,...,L$ of the $d$ conditioned transformations. The mean, variance, and mixture weights of each Gaussian in \eqref{eq:ngnf_inf_base_transformation} are the output of the Neural Network of the corresponding layer. 
For each $m=1,...,d$, the $m$-th Gaussian mixture at level $l$ contains $G_{l,m}>0$ elements. Consequently, the corresponding Neural Network takes an input in dimension $m+d+p+1$ and outputs a $3\times G_{l,m}$ dimensional vector. 
Softmax activation functions ensure that the mixture weights sum to 1. 
In equation form: 
\begin{equation} \label{eq:ngnf_inf_Mean_variance_weights}
    [\boldsymbol{\alpha}^{l,m+1}; \boldsymbol{\mathrm{m}}^{l,m+1};\boldsymbol{\sigma}^{l,m+1}]=\mathrm{NN}^{l,m+1}_{{\theta}^{l,m+1}(\tau)}(\boldsymbol{x}_{1:m}^1,\boldsymbol{x_0},\boldsymbol{\tilde{\mu}}), \quad \text{for $m=0,...,d-1$, and \:$l=2,...,L$},
\end{equation}
where $\mathrm{NN}^{l,m+1}_{{\theta}^{l,m+1}}$ is a Neural Network with parameters $\boldsymbol{\theta}^{l,m+1}$. The architecture of these Neural Networks is based upon GRU cells, which are particularly effective for modeling solutions to PDEs that display sharp transitions in the initial conditions \cite{SIRIGNANO20181339}.
With these ingredients, we define the bijection from $[-1,1]^{d}$ to $[-1,1]^{d}$ applied at layer $l=2,...,L$ and for $m=0,...,d-1$
\begin{equation} \label{eq:ngnf_inf_full_transformation_layer_l}
\begin{split}
     {x}^l_{m+1} & = {n}_{\theta^{l}(\tau)}^{l,m+1}({x}^{l-1}_{m+1}|\boldsymbol{x}^{1}_{1:m},\boldsymbol{x}_{0},\boldsymbol{\tilde{\mu}}) =\tilde{F}^l({x}^{l-1}_{m+1}|\boldsymbol{\theta}^{l,m+1}(\tau), \boldsymbol{x}^{1}_{1:m},\boldsymbol{x}_{0},\boldsymbol{\tilde{\mu}}) \\ & =2[\sum_{k=1}^{G_{l,m+1}} \alpha_{k}^{l,m+1}(\tau)\Phi_{[-1,1]}({x}^{l-1}_{m+1}|\mathrm{m}_{k}^{l,m+1}(\tau),\sigma_{k}^{l,m+1}(\tau))]-1.
\end{split}
\end{equation}
A schematic diagram illustrating the transformation is presented in Appendix \ref{sect:structure_of_the_normalizing_flow}, Figure \ref{fig:ngnf_inference_diagram_flow}.

Crucially, in \eqref{eq:ngnf_inf_full_transformation_layer_l}, ${x}^l_{m+1}$ depends solely on ${x}^{l-1}_{m+1}$ and on $\boldsymbol{x}^{1}_{1:m}$. This implies that the Jacobian matrix of each layer of the map $\boldsymbol{n}_{\boldsymbol{\theta}}$  has a lower triangular structure, which implies that
\begin{equation}\label{eq:ngnf_inf_full_determinant_NF}
\begin{split}
\det(\nabla_{\boldsymbol{x}}\boldsymbol{n}_{\theta(\tau)}(\boldsymbol{x}))
=
\left[\prod_{m=1}^{d}\frac{2}{\hat b_m-\hat a_m}\right]
\prod_{l=2}^{L}\prod_{m=1}^{d}
\left[
2\sum_{k=1}^{G_{l,m}}
\alpha_{k}^{l,m}
\phi_{[-1,1]}\left(x^{l-1}_{m}\mid \mathrm{m}_{k}^{l,m},\sigma_{k}^{l,m}\right)
\right].
\end{split}
\end{equation}

Using \eqref{eq:ngnf_inf_FP_approximation_by_NF} and this structure of our Normalizing Flow, we look for an approximation to the TPDF that solves \eqref{eq:ngnf_inf_TPDF_Fokker_Planck_mu}. In this approximation, the time evolution of the parameter vector $\boldsymbol{\theta}$ is determined by \eqref{eq:ngnf_inf_optimization_Neural_Galerkin}. 
A key feature of the newly introduced Normalizing Flow is that, by appropriately initializing the parameters of the Gaussian mixtures, we can sharply approximate the initial condition of \eqref{eq:ngnf_inf_TPDF_Fokker_Planck_mu} with vanishing-variance truncated Gaussians. 
Moreover, a careful selection of the source distribution $p_Z$ allows us to naturally satisfy homogeneous Dirichlet boundary conditions on $\Gamma=\partial\Omega \cap\partial\hat{\Omega}$. 
In fact, by ensuring that $p_Z(\boldsymbol{n}_{\theta}(\boldsymbol{x}))|_{\Gamma}=0$, the NF will directly meet the desired boundary condition. Instead, rather than directly imposing zero-flux boundary conditions on $\partial\hat{\Omega}\setminus\Gamma$, we minimize the residual \eqref{eq:ngnf_inf_J_cost_functional_Neural_Galerkin} and verify a posteriori that these are satisfied by the obtained numerical solution. We elaborate on these aspects in the Appendix \ref{sect:ngnf_inf_dirac_delta_imposition_zero_flux}.

\section{Bayesian Inference for Fully-Observed Diffusion Processes}

We consider Bayesian inference based on a discretely observed noiseless trajectory $\boldsymbol{y} = [\boldsymbol{y}_{t_0}, \dots, \boldsymbol{y}_{t_n}]$ of the SDE \eqref{eq:ngnf_inf_Diffusion_process_mu}, which is acquired at a constant lag $\Delta$, to infer the parameters of the SDE $\boldsymbol{\mu}$. In this context, the posterior distribution is expressed using Bayes' rule as follows 
    \begin{equation} \label{eq:ngnf_inf_Posterior_params_given_data}
        \pi(\boldsymbol{\mu}|\boldsymbol{y})= \frac{\mathrm{L}(\boldsymbol{\mu}|\boldsymbol{y}){\pi}_0(\boldsymbol{\mu})}{\mathrm{e}(\boldsymbol{y})},
    \end{equation}
    where $\mathrm{L}$ is the likelihood function, $\pi_0$ is the prior distribution, and $\mathrm{e}(\boldsymbol{y})=\int_{\Xi} \mathrm{L}(\boldsymbol{\mu}|\boldsymbol{y}){\pi}_0(\boldsymbol{\mu})d\boldsymbol{\mu}$ is the evidence, usually not explicitly known. Thanks to the Markov property we can factorize the likelihood as follows
    \begin{equation} \label{eq:ngnf_inf_Factor_likelihood}
        \mathrm{L}(\boldsymbol{\mu}|\boldsymbol{y}) = \prod_{i=1}^{n} \rho(\boldsymbol{y}_{t_{i}}|t_i,t_{i-1},\boldsymbol{y}_{t_{i-1}},\boldsymbol{\mu}).
    \end{equation}
 
    By using the approximation $\rho(\boldsymbol{y}_{t_{i}}|t_i,t_{i-1},\boldsymbol{y}_{t_{i-1}},\boldsymbol{\mu})\simeq \Rho(\boldsymbol{y}_{t_{i}}|\boldsymbol{\theta}(\Delta), \boldsymbol{y}_{t_{i-1}},\boldsymbol{\tilde{\mu}})$, the surrogate transition density function proposed in this manuscript provides a positive estimator of the likelihood \eqref{eq:ngnf_inf_Factor_likelihood}. Importantly, our surrogate functions as a valid probability density; thus, it can be directly utilized to conduct Bayesian inference.

    Our method relies on an offline training phase during which the time-dependent parameters of the Normalizing Flow $\boldsymbol{\theta}(\tau)$ are learned by forcing $\Rho$ to satisfy the Fokker-Planck equation \eqref{eq:ngnf_inf_TPDF_Fokker_Planck_mu} for any $\boldsymbol{\tilde{\mu}}=[\boldsymbol{\mu},s]$ of interest. Learning $\boldsymbol{\theta}(\tau)$ for $\tau\in[0,\Delta]$, rather than only at the final time $\tau=\Delta$, allows access to a time-dependent set of parameters. This enables inference even when the observations $\boldsymbol{y}$ are collected asynchronously, potentially at a higher rate than $\Delta$, thereby providing greater flexibility in real-time applications.
    During inference, we can amortize the computational costs incurred during training. Estimating \eqref{eq:ngnf_inf_Factor_likelihood} requires merely applying the change of variable formula \eqref{eq:ngnf_inf_FP_approximation_by_NF} to each of the $n$ observations $\{\boldsymbol{y}_{t_i}\}_{i=1}^{n}$, while conditioning on the previous observations $\{\boldsymbol{y}_{t_{j-1}}\}_{j=1}^{i}$ and the current state of the chain targeting \eqref{eq:ngnf_inf_Posterior_params_given_data}.
     In the following experiments, we generate the data $\boldsymbol{y}$ by simulating the SV model with an unknown parameter $\boldsymbol{\mu}^{\star}\in\Xi$. Inference consists in sampling from the posterior distribution $\pi$ given the data.

\section{Numerical Results} \label{sect:ngnf_inf_Numerical_results}
In this section, we perform Bayesian inference for the Heston stochastic volatility model and for the SVCEV model, a non-affine extension of Heston. 
We assess the performance of our model in likelihood estimation by comparing it to competing methods based either on 
 Aït-Sahalia's closed-form likelihood expansion \cite{AITSAHALIA2007413}, or 
the delta expansion of the TPDF based on the Itô-Taylor
expansion proposed in \cite{YANG2019256}.
An additional experiment performed in a higher-dimensional setting is included in Appendix \ref{sect:ngnf_inf_BDFS_subsection}. 
The details of the data generation process, model's setup and training can be found in Appendix \ref{sect:ngnf_inf_Heston_subsection_supplementary}, \ref{sect:ngnf_inf_svcev_subsection_supplementary}.
\subsection{Heston model} \label{sect:ngnf_inf_Heston_subsection}

The Heston model is a two-dimensional stochastic volatility model which describes the dynamics of the log price of a stock or of the short term interest rate, whose variance is defined by the CIR process \cite{384845b8-4a63-3ff6-9340-656650925d85}. The model can be expressed as
\begin{equation} \label{eq:ngnf_inf_Heston_model}
    \begin{cases}
        dV(t) = \beta(\alpha-V(t))dt + \sqrt{V(t)}\sigma dW(t) \\
        dY(t) = (\mu - \frac{V(t)}{2} ) dt + \sqrt{V(t)}(\rho dW(t) + \sqrt{1-\rho^2} dB(t)),
    \end{cases}
\end{equation}
where $W(t)$ and $B(t)$ are independent Brownian motions. 
The Normalizing Flow is parametrized by the location of the initial condition $[V(0),Y(0)]=\boldsymbol{x_0}\in\mathbb{R}^+\times\mathbb{R}$ and by the parameters of the SDE $\boldsymbol{\mu}=[\alpha,\beta,\sigma,\mu,\rho]\in\mathbb{R}^+\times\mathbb{R}^+\times\mathbb{R}^+\times\mathbb{R}\times(-1,1)$.
The residual \eqref{eq:ngnf_inf_J_cost_functional_Neural_Galerkin} is minimized over a $5+2+2$ dimensional space, which makes the problem computationally challenging for traditional solvers.

\begin{figure}[t]
\centering
\includegraphics[width=\linewidth]{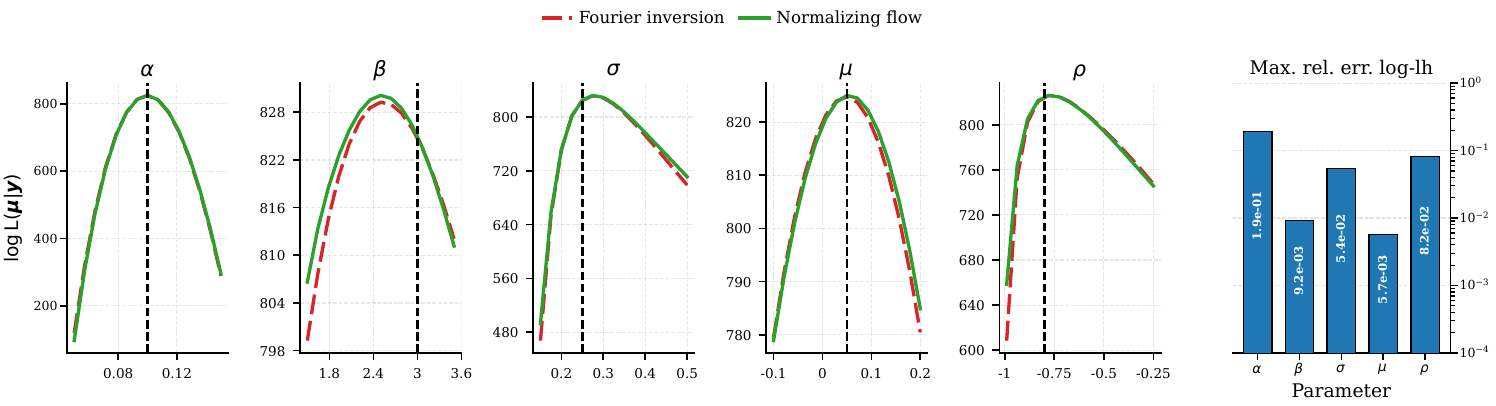}
\caption{Log-likelihood on Heston-generated data, with error bars showing the maximum relative error for each 1D slice. 
Unshown parameters are fixed at $\boldsymbol{\mu}^{\star}$; the vertical black line marks the true parameter.}
\label{fig:ngnf_inf_NG_vs_Fourier_log_lh_heston}
\end{figure}

For this model, the exact characteristic function, which is the Fourier transform of the TPDF, is available in closed form \cite{Lamoureux}. We use this example to quantitatively assess the performance of our method in estimating the true likelihood, obtained via Fourier inversion of the TPDF. To accomplish this, we generate $n+1=350$ synthetic observations by simulating a single trajectory of the Heston model \eqref{eq:ngnf_inf_Heston_model}, using the parameters $\boldsymbol{\mu}^{\star} = [0.1, 3, 0.25, 0.05, -0.8]$ and $\Delta = 0.5$.
In Figure \ref{fig:ngnf_inf_NG_vs_Fourier_log_lh_heston}, we compare the approximation of the log-likelihood (log-lh) \eqref{eq:ngnf_inf_Factor_likelihood} evaluated over the observations conditioned on the true parameter \(\boldsymbol{\mu}^{\star}\).
The Normalizing Flow is capable of providing an approximation that reflects the behavior of the semi-exact likelihood, making it a suitable candidate for an MCMC algorithm targeting the posterior. From a quantitative perspective, we assess the quality of the surrogate transition density in three complementary ways. First, we report the relative error between the reference log-likelihood and the NF-based log-likelihood at the true parameter $\boldsymbol{\mu}^{\star}$.
Second, we draw a batch of conditioning pairs $\{(\boldsymbol{x}_{0,i},\boldsymbol{\mu}_i)\}_{i=1}^{N_{\mathrm{test}}}\sim\eta\times\hat{\pi}$ (cf. \eqref{test_heston_distribution}), with $N_{\mathrm{test}}=100$. For each pair, we compute the relative $L^2(\Omega)$ error at time $\Delta$ between the reference TPDF and its NF approximation.
Third, to specifically probe the challenging boundary regime of the volatility process, we report the same $L^2(\Omega)$ error on a low-volatility test case with fixed initial variance $v_0 = \varepsilon$, where $\varepsilon \ll 1$. 
The results are summarized in Table~\ref{tab:heston_quantitative_error} and suggest that the NF generalizes properly across different parameters and is resilient in low volatility regimes.

\begin{table}[t]
\centering
\caption{Quantitative accuracy of the NF surrogate for the Heston model \eqref{eq:ngnf_inf_Heston_model}}
\label{tab:heston_quantitative_error}
\begin{tabular}{lc}
\toprule
Metric & Value \\
\midrule
Relative error of the log-lh \eqref{eq:ngnf_inf_Factor_likelihood} at $\boldsymbol{\mu}^{\star}$ & 0.0003  \\
Mean (std. error) relative $L^2(\Omega)$ error over $\{\boldsymbol{x}_{0,i},\boldsymbol{\mu}_i\}_{i=1}^{N_{\mathrm{test}}}$ & 0.0633 (0.0123) \\
Relative $L^2(\Omega)$ error for $(v_0=\epsilon,y_0,\boldsymbol{\mu}^{\star})$ & 0.04 ($\epsilon=10^{-2}$),  0.06 ($\epsilon=10^{-4}$)\\ 
\bottomrule
\end{tabular}
\end{table}

We use the semi-exact TPDF and the surrogate model to draw samples from the posterior distribution, with MCMC. 
To achieve this, we use MCMC in the form of Slice Sampling with axis-aligned \cite{Neal2003SliceSampling}.
In Table \ref{tab:runtime_comparison_heston}, we report the empirical Wasserstein-2 ($W_2$) distance \cite{Villani2008OptimalTransport} between the reference posterior distribution and the posterior distribution drawn using NF, along with the computational runtime required to sample the chain. 
Importantly, our model remains comparable to  Fourier transform inversion while being significantly more cost-effective and maintains high accuracy. Our approximation is much faster because it requires only $n$ forward passes through the Flow, rather than approximating the Fourier inversion. 
In Appendix \ref{sect:additional_illustrations_heston}, Figures \ref{fig:ngnf_inference_Posterior_heston_marginal}, \ref{fig:ngnf_inf_NG_posterior_heston} display the marginal posterior distributions as well as the joint distribution of two out of the five parameters of the Heston model.  The joint distributions for $\beta - \sigma$ and $\rho - \sigma$ show that our model effectively captures the correlation between different parameters. These results further suggest that our model is approaching the correct posterior distribution.
\begin{table}[t]
\centering
\caption{Runtime for $10^4$ MCMC samples, empirical joint $W_2$ distance between sampled posterior distributions $\pi(\boldsymbol{\mu}|\boldsymbol{y})$, and marginal parameter-wise $W_2$ distances, relative to the Fourier reference.}
\label{tab:runtime_comparison_heston}
\begin{tabular}{llccccccc}
\toprule
Model & Method & Time (s) & $W_2$ & $W_2^\alpha$ & $W_2^\beta$ & $W_2^\sigma$ & $W_2^\mu$ & $W_2^\rho$ \\
\midrule
\multirow{2}{*}{Heston \eqref{eq:ngnf_inf_Heston_model}}
& Fourier & 14560.13 & Ref. & -- & -- & -- & -- & -- \\
& \textbf{NF} & 2243.37 & 0.066 & 0.0003 & 0.057 & 0.0036 & 0.0046 & 0.011 \\
\bottomrule
\end{tabular}
\end{table}

Finally, we generate $n+1=200$ data in a low-frequency setting, using a time lag $\Delta=1$, $\boldsymbol{\mu}^{\star}$ as defined earlier, and compare various likelihood approximations.
In Table~\ref{tab:heston_compare_quantitative_error_models_delta_1}, we report the relative error between the reference log-likelihood and the approximated log-likelihood at the true parameter $\boldsymbol{\mu}^{\star}$ and averaged over a batch $\{\boldsymbol{\mu}^{i}\}_{i=1}^{N_{\mathrm{test}}}\sim\hat{\pi}$ (cf. \eqref{test_heston_distribution}) with $N_{\mathrm{test}}=100$. 
We consider both NF and the 2nd order closed-form expansion from Aït-Sahalia (A.-S.) \cite{ait_sahalia2008closed} and the 4th order It\^o-Taylor expansion \cite{YANG2019256}.
Among the approximations evaluated, NF attains the smallest trajectory log-likelihood errors. Notably, our approach is particularly effective in the low-frequency setting and away from $\boldsymbol{\mu}^{\star}$ because it does not rely on expansions in powers of $\Delta$ but rather a structure-preserving surrogate for the TPDF.
\begin{table}[t]
\centering
\caption{Accuracy of the likelihood approximations for the Heston model \eqref{eq:ngnf_inf_Heston_model}.  
Best values in bold.}
\label{tab:heston_compare_quantitative_error_models_delta_1}
\begin{tabular}{lccc}
\toprule
Metric & NF & A.-S. & It\^{o}-Taylor \\
\midrule
Relative error over the log-lh \eqref{eq:ngnf_inf_Factor_likelihood} given $\boldsymbol{\mu}^{\star}$ 
& \textbf{0.0044} & 0.1934 & 0.0441 \\
\midrule
Average relative err. of log-lh over $\{\boldsymbol{\mu}^{i}\}_{i=1}^{N_{\mathrm{test}}}$ 
& $\boldsymbol{0.257}$ & 4.335 & 3.232 \\
Standard error (relative err. of log-lh) 
& $0.039$ & 0.714 & 0.539 \\
Median relative err. of log-lh over $\{\boldsymbol{\mu}^{i}\}_{i=1}^{N_{\mathrm{test}}}$ 
& \textbf{0.100} & 3.064 & 1.350 \\
\bottomrule
\end{tabular}
\end{table}

\subsection{SVCEV}\label{sect:ngnf_inf_svcev}

We now consider the SVCEV model, which is nonlinear, irreducible, and non-affine, and therefore its characteristic function cannot be expressed in closed-form:
\begin{equation} \label{eq:ngnf_inf_SVCEV}
    \begin{cases}
        dV(t) = \beta(\alpha-V(t))dt + \sigma V^{\gamma}(t)dW(t) \\
        dY(t)=(\mu-V(t)/2)dt + \sqrt{V(t)}(\rho dW(t)+\sqrt{1-\rho^2} dB(t)).
    \end{cases}
\end{equation}
 We generate $n+1=100$ data with time-lag $\Delta=0.5$, and compare our results with a reference estimator based on data augmentation (DA) and importance sampling \cite{Durham01072002}, see Appendix \ref{sect:ngnf_inf_svcev_subsection_supplementary} for further details.
 In Table~\ref{tab:svcev_quantitative_error}, we analyze the relative error between the reference and the approximated log-likelihood at the true parameter $\boldsymbol{\mu}^{\star}$ and averaged over a batch $\{\boldsymbol{\mu}^{i}\}_{i=1}^{N_{\mathrm{test}}}\sim\hat{\pi}$ (cf. \eqref{test_svcev_distribution}) with $N_{\mathrm{test}}=100$. 
 We compare our method (NF), the 2nd order closed-form approximation of A.-S. \cite{ait_sahalia2008closed}, and the 4th order It\^o-Taylor expansion \cite{YANG2019256}.
 From this table, we can conclude that the true error is upper-bounded by the sum of the discrepancy with respect to the DA reference and the DA estimator's error.
 NF yields the smallest discrepancies, even away from $\boldsymbol{\mu}^{\star}$, underscoring the benefit of learning the TPDF over $\tilde{\pi}$ for Bayesian inference. By contrast, we observed that closed-form transition-density expansions can become unstable and, in some cases, diverge when evaluated far from $\boldsymbol{\mu}^{\star}$. 

\begin{figure}[H]
\centering
\includegraphics[width=\textwidth]{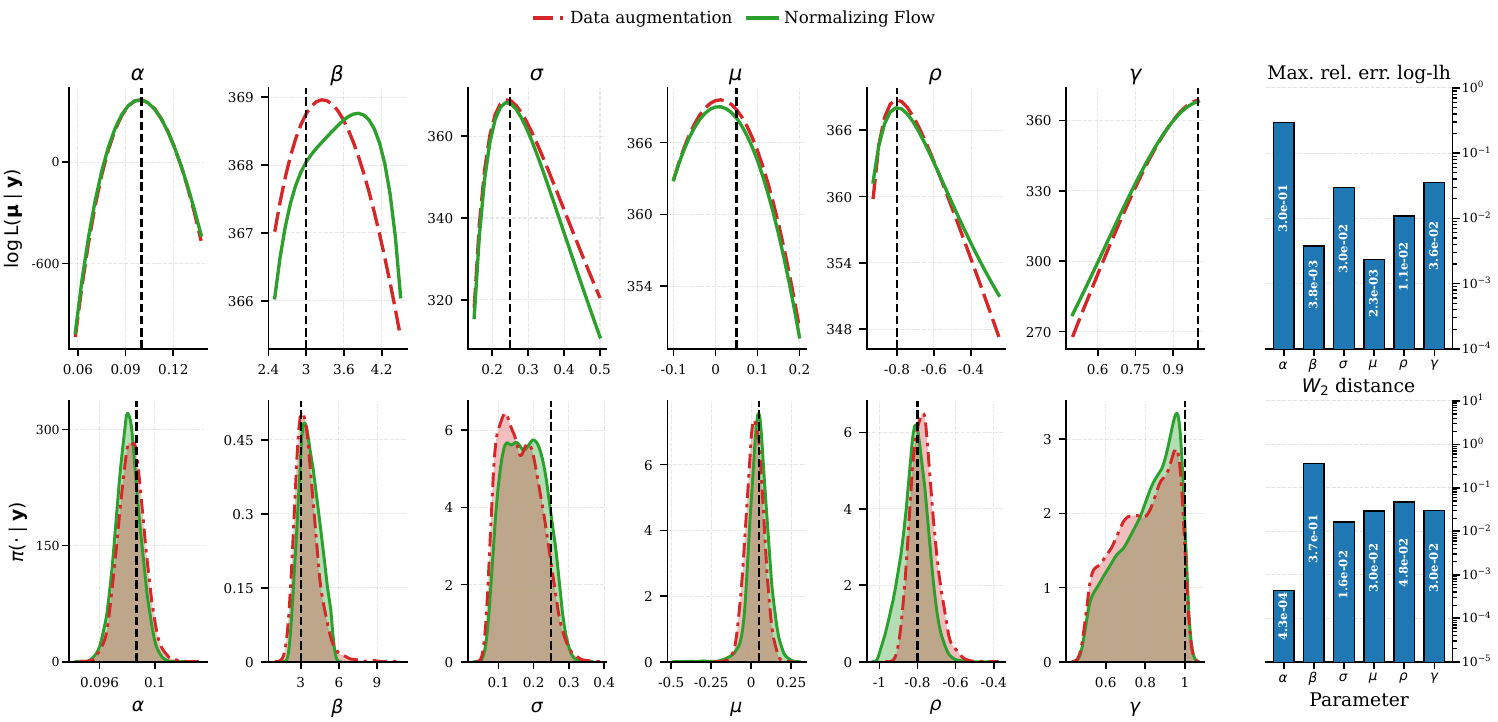}
\caption{Top: log-lh slices for SVCEV data, with bars of the maximum relative-error of log-lh  and all other parameters fixed at $\boldsymbol{\mu}^{\star}$. Bottom: posterior marginals with corresponding $W_2$ distances.}
\label{fig:ngnf_inference_NG_vs_data_augmentation_log_lh_and_post_svcev}
\end{figure}

\begin{table}[t]
\centering
\caption{Relative log-likelihood discrepancies for the SVCEV model \eqref{eq:ngnf_inf_SVCEV}. Best values in bold. At $\boldsymbol{\mu}^{\star}$, the DA estimator itself has estimated relative error $5.5\times 10^{-4}$ with 95\% CI $[4.96,\,5.96]\times 10^{-4}$.}
\label{tab:svcev_quantitative_error}
\begin{tabular}{lccc}
\toprule
Metric & NF & A.-S. & Itô-Taylor \\
\midrule
Relative err. of the log-lh \eqref{eq:ngnf_inf_Factor_likelihood} at $\boldsymbol{\mu}^{\star}$ 
& \textbf{0.0004}  & 0.0063  & 0.0091 \\
\midrule
Average relative err. of log-lh over $\{\boldsymbol{\mu}^{i}\}_{i=1}^{N_{\mathrm{test}}}$ 
& \textbf{0.084} & 0.941 & 0.811 \\
Standard error (relative err. of log-lh) 
& {0.014} & 0.163 & 0.112 \\
\bottomrule
\end{tabular}
\end{table}
Lastly, in Figure \ref{fig:ngnf_inference_NG_vs_data_augmentation_log_lh_and_post_svcev}, we evaluate the quality of the log-likelihood and posterior distribution obtained by the Normalizing Flow against the data augmentation reference.  
The results show good agreement, indicating that the surrogate Normalizing Flow extends its applicability beyond the affine Heston framework. 
 Additionally, as shown in Table \ref{tab:runtime_comparison_svcev}, we find that the Normalizing Flow yields a substantial computational speed-up, making it a more efficient alternative to traditional inference methodologies.

\begin{table}[t]
\centering
\caption{SVCEV model \eqref{eq:ngnf_inf_SVCEV}: runtime for $10^4$ MCMC samples and empirical $W_2$ distance relative to the DA reference. A.-S. is omitted as it does not target the posterior.}
\label{tab:runtime_comparison_svcev}
\begin{tabular}{lccc}
\toprule
Metric & Data augmentation & Normalizing Flow & It\^o Taylor \\
\midrule
Time (s) & 63541.69 & \textbf{1389.25} & 7219.16 \\
$W_2$ distance & Reference & \textbf{0.412} & 1.337 \\
\bottomrule
\end{tabular}
\end{table}

\section{Future directions and limitations}\label{sect:ngnf_inf_conclusions}

In this work, we have proposed a surrogate model based on Neural Galerkin and Normalizing Flows for likelihood estimation and Bayesian inference of diffusion processes.
Compared with traditional inference techniques, NFs offer significant accuracy and computational advantages, revealing a powerful tool for exploring high-dimensional posterior distributions of SDE parameters.
We focused on fully observed processes; nevertheless, our model can be combined with particle-based MCMC methods to infer partially observed diffusions, a direction we leave for future research.

A key assumption of our work is that the target density can be effectively modeled using a distribution with compact support. 
The model's accuracy is not guaranteed in the tails of the distributions, which makes it unsuitable for directly simulating extreme events. Likewise, the reference distribution of the NF may not accurately represent the tail decay in the inaccessible boundary. A possible remedy is to parameterize and train this reference density, which we will explore in future research.
In this extension, our approach would be well-suited for modeling processes living in bounded domains, such as those associated with Fisher-Wright type SDEs with inaccessible boundaries. 
Finally, future research will explore our methodology and the generative potential of Normalizing Flows as a promising route for modeling and inference of  McKean-Vlasov SDEs.

\bibliography{biblist}  

\clearpage
\appendix

\renewcommand\thepart{}
\renewcommand\partname{}

\section{Structure of the Normalizing Flow}\label{sect:structure_of_the_normalizing_flow}

\begin{figure}[!htb]
  \centering
  \includegraphics[width=\linewidth]{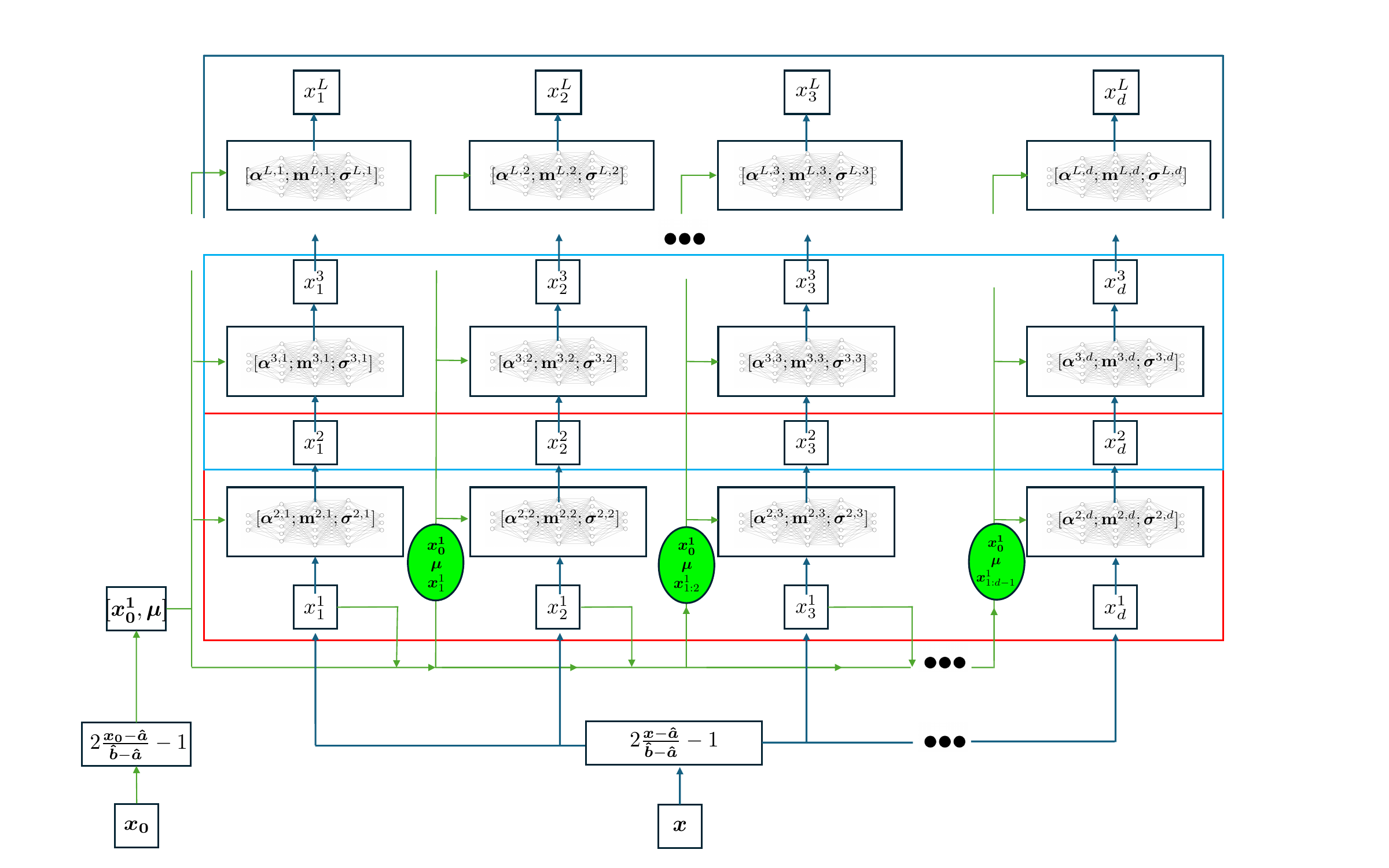}
  \caption{Transformation executed by the Normalizing Flow \eqref{eq:ngnf_inf_NF_mapping} across layers $l=1,...,L$.
Moving from bottom to top, each column $m=1,...,d$ transforms $x_{m}$ while conditioning on $\boldsymbol{x_{1:m-1}^1}$, $\boldsymbol{x_0}$, $\boldsymbol{\mu}$.}
  \label{fig:ngnf_inference_diagram_flow}
\end{figure}

The multi-layer schematic of the bounded Normalizing Flow is shown in Figure \ref{fig:ngnf_inference_diagram_flow}.

\section{Approximation of the Dirac delta and zero-flux boundary conditions}\label{sect:ngnf_inf_dirac_delta_imposition_zero_flux}

\subsection{Imposition of the Dirac delta initial condition}

    The bounded Normalizing Flow enables us to utilize the Jacobian of the determinant from the transformation \eqref{eq:ngnf_inf_full_determinant_NF} for representing the initial condition of the Fokker-Planck PDE. To do this, we initialize weights and biases of the output layer of the $d$ Neural Networks of the second set of bijection $ {n}_{\theta^{2}(\tau)}^{2,m+1}$, for $m=0,...,d-1$, to zero. Specifically, only the entries of $\boldsymbol{\theta}^{2}(0)$ that compose the weights and biases of the last layer of the Neural Network are initialized to zero. The variances of the Gaussian distributions in the mixture of this layer are set to an arbitrary tolerance $\hat{\epsilon}>0$, which approaches zero. Simultaneously, we adjust the means of these Gaussian distributions to coincide with the location of the initial condition, $\boldsymbol{x_0}$.
    The weights of the Gaussians are initialized uniformly.
    This initialization results in mixtures of truncated Gaussians with the following parameters 
    \begin{equation}
        \begin{cases}
            \begin{array}{ll}
                {\mathrm{m}}_k^{2,m+1}={x_{0,m+1}}, & \text{for $m=0,...,d-1$, $k=1,...,G_{2,m+1}$,} \\
                 {\sigma}_k^{2,m+1}=\hat{\epsilon} & \text{for $m=0,...,d-1$, $k=1,...,G_{2,m+1}$,} \\
                 \alpha_k^{2,m+1}=1/G_{2,m+1} & \text{for $m=0,...,d-1$, $k=1,...,G_{2,m+1}$.} \\
            \end{array}
        \end{cases}
    \end{equation}
    As a result, the transformation collapses to a step function, contributing to the determinant as "$\frac{\partial n^{2,m}_{{\theta}^{l}(\tau)}({x}^{1}_m)}{\partial x^{1}_m}\simeq\delta_{x_0}((\boldsymbol{x^0})_m)$" for $m=1,...,d$.  These transformations map all the elements satisfying the componentwise inequality $\boldsymbol{x}\geq\boldsymbol{x_0}$ to $\boldsymbol{1}$ and the remaining elements to $\boldsymbol{-1}$. 
    The weights of the Neural Networks in the subsequent layers of the Flow are initialized randomly, and positive variance is ensured by applying the softplus activation function. Consequently, compared to the contribution from the determinant of the Jacobian of the transformation in the second layer, the contributions to $\mathrm{det}(\nabla_{\boldsymbol{x}}\boldsymbol{n}_{\theta})$ from the  following layers (for $l\geq 3$) are negligible.
     The resulting density is not negligible only in a neighborhood of $\boldsymbol{x_0}$, and its sharpness is calibrated  heuristically and controlled by $\hat{\epsilon}$. 

      We further remark that the strength of the Flow's structure based on Gaussian mixture models lies in its ability to approximate the Dirac delta distribution with only the first trainable layer.

\subsection{Zero-flux boundary conditions}
 
When we truncate the domain \(\Omega\) into \(\hat{\Omega}\), we theoretically impose zero-flux boundary conditions on \(\partial\hat{\Omega}\setminus\Gamma\). For the diffusion process \eqref{eq:ngnf_inf_Diffusion_process_mu}, these conditions are expressed as follows:
\begin{equation}\label{eq:zero_flux_bc}
\varphi(\boldsymbol{x},t,s,\boldsymbol{x_0},\boldsymbol{\mu})=\boldsymbol{J}(\boldsymbol{x},t,s,\boldsymbol{x_0},\boldsymbol{\mu})\cdot\boldsymbol{n}=0\quad \text{for all $\boldsymbol{x}\in\partial\hat{\Omega}\setminus\Gamma$},
\end{equation}
where $\boldsymbol{n}$ is the outer normal unit vector at $\partial\hat{\Omega}\setminus\Gamma$ and
\begin{equation}
    \boldsymbol{J}(\boldsymbol{x},t,s,\boldsymbol{x_0},\boldsymbol{\mu})= \boldsymbol{b}(t,\boldsymbol{x},\boldsymbol{\mu})\rho(\boldsymbol{x}|t,s,\boldsymbol{x_0},\boldsymbol{\mu})-\frac{1}{2}\nabla_{\boldsymbol{x}}\cdot(\Sigma(t,\boldsymbol{x},\boldsymbol{\mu})\rho(\boldsymbol{x}|t,s,\boldsymbol{x_0},\boldsymbol{\mu})).
\end{equation}
In our experiments, instead of imposing the boundary condition given in \eqref{eq:zero_flux_bc}, we verify that these conditions are satisfied by the numerical solution. This approach is justified by the expectation that, since we initialize the dynamics from a Dirac distribution, most of the mass of the TPDF will be contained within \(\hat{\Omega}\), provided that \(\hat{\Omega}\) is sufficiently large.
To support our statement, we verify that our solution satisfies 
\begin{equation}\label{eq:ngnf_inf_zero_flux_bc}
\int_s^{s+\Delta}\int_{\partial\hat{\Omega}\setminus\Gamma}\lvert \boldsymbol{J}(\boldsymbol{x},t,s,\boldsymbol{x_0},\boldsymbol{\mu})\cdot\boldsymbol{n}\rvert d\boldsymbol{x}dt\simeq0.
\end{equation}
We validate this observation for both the Heston model \eqref{eq:ngnf_inf_Heston_model} and the SVCEV model \eqref{eq:ngnf_inf_SVCEV}. For each model, we sample \( N_{\mathrm{test}} = 100 \) conditioning pairs \( \{(\boldsymbol{x}_{0,i},\boldsymbol{\mu}_i)\}_{i=1}^{N_{\mathrm{test}}} \sim \eta \times \tilde{\pi}_0 \). For each pair, we use our Normalizing Flow to approximate \eqref{eq:zero_flux_bc} and apply the trapezoidal rule to integrate \eqref{eq:ngnf_inf_zero_flux_bc} in both time and space.
Table~\ref{tab:zero_flux_verification} reports the empirical mean, standard deviation, and 95\% confidence interval of the numerical approximation of \eqref{eq:ngnf_inf_zero_flux_bc} and of the maximum absolute flux 
\begin{equation}\label{eq:ngnf_inf_max_flux_bc}
\max_{(t,\boldsymbol{x})\in [s,s+\Delta]\times\partial\hat{\Omega}\setminus\Gamma}
\left|\boldsymbol{J}(\boldsymbol{x},t,s,\boldsymbol{x}_0,\boldsymbol{\mu})\cdot\boldsymbol{n}\right|.
\end{equation}
\begin{table}[t]
\centering
\caption{
A posteriori verification of the zero-flux boundary condition on 
$\partial\hat{\Omega}\setminus\Gamma$ over $N_{\mathrm{test}}=100$ test pairs $ \{(\boldsymbol{x}_{0,i},\boldsymbol{\mu}_i)\}_{i=1}^{N_{\mathrm{test}}} \sim \eta \times \tilde{\pi}_0 $.}
\label{tab:zero_flux_verification}
\begin{tabular}{llccc}
\toprule
Model & Quantity & Mean & Std. dev. & 95\% CI for mean \\
\midrule
\multirow{2}{*}{Heston} 
& Time-integrated flux \eqref{eq:ngnf_inf_zero_flux_bc} 
& $3.08{\times}10^{-11}$ 
& $7.71{\times}10^{-11}$ 
& $[2.01,4.15]{\times}10^{-11}$ \\

& Maximum absolute flux \eqref{eq:ngnf_inf_max_flux_bc}
& $2.34{\times}10^{-8}$
& $1.36{\times}10^{-7}$ 
& $[3.97,5.52]{\times}10^{-8}$ \\

\midrule
\multirow{2}{*}{SVCEV} 
& Time-integrated flux \eqref{eq:ngnf_inf_zero_flux_bc}
& $5.87{\times}10^{-10}$ 
& $3.05{\times}10^{-9}$ 
& $[1.64,10.09]{\times}10^{-10}$ \\

& Maximum absolute flux \eqref{eq:ngnf_inf_max_flux_bc}
& $2.40{\times}10^{-7}$
& $1.43{\times}10^{-6}$ 
& $[2.54,56.29]{\times}10^{-8}$ \\
\bottomrule
\end{tabular}
\end{table}

\subsubsection{Imposition of the zero-flux boundary conditions in the Neural Galerkin framework}
As an alternative, we propose an approach to impose zero-flux boundary conditions while integrating the dynamics \eqref{eq:ngnf_inf_optimization_Neural_Galerkin}. 
Rather than minimizing \eqref{eq:ngnf_inf_J_cost_functional_Neural_Galerkin} alone, one could tackle the minimization problem 
\begin{equation} \label{eq:ngnf_inf_optimization_Neural_Galerkin_constrained}
\begin{split}
    \min_{\boldsymbol{\zeta}\in\dot{\Theta}} &\int_{\Xi^{\star}}\int_{\Omega}\int_{\Omega} \lvert r_{s+\tau,s}(\boldsymbol{\theta},\boldsymbol{\zeta},\boldsymbol{x},\boldsymbol{x_0},\boldsymbol{\tilde{\mu}}) \rvert ^{2} d\nu_{{\theta}(\tau)}(\boldsymbol{x}|\boldsymbol{x_0},\boldsymbol{\tilde{\mu}}) d\eta(\boldsymbol{x_0})d\tilde{\pi}_0^{\star}(\boldsymbol{\tilde{\mu}})\\ &+\epsilon\int_{\Xi^{\star}}\int_{\Omega}\int_{\partial\hat{\Omega}\setminus\Gamma} \lvert \nabla_{\boldsymbol{\theta}}\Rho(\boldsymbol{x}|\boldsymbol{\theta}(\tau),\boldsymbol{x_0},\boldsymbol{\tilde{\mu}})\boldsymbol{\zeta}-\frac{1}{\epsilon}\varphi(\boldsymbol{x},t,s,\boldsymbol{x_0},\boldsymbol{\mu}) \rvert ^{2} d\tilde{\nu}(\boldsymbol{x}|\boldsymbol{x_0},\boldsymbol{\tilde{\mu}}) d\eta(\boldsymbol{x_0})d\tilde{\pi}_0^{\star}(\boldsymbol{\tilde{\mu}}), 
\end{split}
\end{equation}
where $\epsilon>0$ is a regularization parameter and $\tilde{\nu}$ is a positive measure over $\partial\hat{\Omega}\setminus\Gamma$. 
The minimizers of \eqref{eq:ngnf_inf_optimization_Neural_Galerkin_constrained} are given by the Euler-Lagrange equations 
\begin{equation}\label{eq:ngnf_inf_zero_flux_euler_lagrange}
    (M^{i}(\boldsymbol{\theta})+\epsilon M^{b}(\boldsymbol{\theta}))\boldsymbol{\zeta} = \boldsymbol{F}^{i}(t,\boldsymbol{\theta})+ \boldsymbol{F}^{b}(t,\boldsymbol{\theta}),
\end{equation}
where the internal contributions read
\begin{equation}
\begin{aligned}
    M^{i}(\boldsymbol{\theta})
    &=
    \int_{\Xi^{\star}}\int_{\Omega}\int_{\Omega}
    \nabla_{\boldsymbol{\theta}}\Rho(\boldsymbol{x}|\boldsymbol{\theta}(\tau),\boldsymbol{x}_0,\tilde{\boldsymbol{\mu}})
    \nabla_{\boldsymbol{\theta}}\Rho(\boldsymbol{x}|\boldsymbol{\theta}(\tau),\boldsymbol{x}_0,\tilde{\boldsymbol{\mu}})^T
    \,d\nu_{\boldsymbol{\theta}(\tau)}
    \,d\eta 
    \,d\tilde{\pi}_0^{\star},
    \\[0.5em]
    \boldsymbol{F}^{i}(t,\boldsymbol{\theta})
    &=
    \int_{\Xi^{\star}}\int_{\Omega}\int_{\Omega}
    \nabla_{\boldsymbol{\theta}}\Rho(\boldsymbol{x}|\boldsymbol{\theta}(\tau),\boldsymbol{x}_0,\tilde{\boldsymbol{\mu}})
    \mathcal{L}^{\star}_t
    \bigl(\Rho(\cdot|\boldsymbol{\theta}(\tau),\boldsymbol{x}_0,\tilde{\boldsymbol{\mu}})\bigr)(\boldsymbol{x})
    \,d\nu_{\boldsymbol{\theta}(\tau)}
    \,d\eta
    \,d\tilde{\pi}_0^{\star},
\end{aligned}
\end{equation}
while the boundary contributions are given by
\begin{equation}
\begin{aligned}
    M^{b}(\boldsymbol{\theta})
    &=
    \int_{\Xi^{\star}}\int_{\Omega}\int_{\partial\hat{\Omega}\setminus\Gamma}
    \nabla_{\boldsymbol{\theta}}\Rho(\boldsymbol{x}|\boldsymbol{\theta}(\tau),\boldsymbol{x}_0,\tilde{\boldsymbol{\mu}})
    \nabla_{\boldsymbol{\theta}}\Rho(\boldsymbol{x}|\boldsymbol{\theta}(\tau),\boldsymbol{x}_0,\tilde{\boldsymbol{\mu}})^T
    \,d\tilde{\nu}
    \,d\eta
    \,d\tilde{\pi}_0^{\star},
    \\[0.5em]
    \boldsymbol{F}^{b}(t,\boldsymbol{\theta})
    &=
\int_{\Xi^{\star}}\int_{\Omega}\int_{\partial\hat{\Omega}\setminus\Gamma}
    \nabla_{\boldsymbol{\theta}}\Rho(\boldsymbol{x}|\boldsymbol{\theta}(\tau),\boldsymbol{x}_0,\tilde{\boldsymbol{\mu}})
    \varphi(\boldsymbol{x},t,s,\boldsymbol{x}_0,{\boldsymbol{\mu}})
    \,d\tilde{\nu}
    \,d\eta
    \,d\tilde{\pi}_0^{\star}.
\end{aligned}
\end{equation}
It can be shown that, as $\epsilon\rightarrow0$, \eqref{eq:ngnf_inf_zero_flux_euler_lagrange} coincides with the weak formulation of the Fokker-Planck equation \eqref{eq:ngnf_inf_TPDF_Fokker_Planck_mu} on the bounded domain $\hat{\Omega}$ with zero-flux boundary conditions on $\partial\hat{\Omega}\setminus\Gamma$ and homogeneous Dirichlet data on $\Gamma$, taking as test function $\nabla_{\boldsymbol{\theta}}\Rho$.
We plan to explore this alternative approach for imposing zero-flux boundary conditions in future studies.

\section{Setup and further validation for the Heston Stochastic Volatility Model}\label{sect:ngnf_inf_Heston_subsection_supplementary}

To generate the data for the test case, we use the Euler-Maruyama scheme to simulate a trajectory of $n+1=350$ observations with parameter values of $\alpha^{\star}=0.1$, $\beta^{\star}=3$, $\sigma^{\star}=0.25$, $\mu^{\star}=0.05$, and $\rho^{\star}=-0.8$. The observation gap is set to $\Delta=0.5$. 
To minimize discretization errors, as outlined in references \cite{AITSAHALIA2007413, StramerBognar2011}, we introduce a discretization made of 100 sub-intervals per sampling interval. We generate a total of 700 observations and then drop the first 350. These last 350 observations constitute our data $\boldsymbol{y} = [\boldsymbol{y}_{t_0}, \dots, \boldsymbol{y}_{t_{349}}]$. 

As the SDE in equation \eqref{eq:ngnf_inf_Heston_model} is autonomous, we can set the initial time for integration to an arbitrary value, such as \(s=0\). Thus, the dynamics can be integrated from that time onward.
Let $\boldsymbol{x}=[v,y]$ and let us denote  $\rho(\boldsymbol{x}|t)=\rho(\boldsymbol{x}|t,0,\boldsymbol{x_0},\boldsymbol{\mu})$,
The Fokker-Planck equation associated with \eqref{eq:ngnf_inf_Heston_model} reads
\begin{equation} \label{FP_Heston}
    \begin{cases}
    \begin{array}{ll}
        \partial_t{\rho(\boldsymbol{x}|t)} + (\mu - v/2)\partial_y{\rho(\boldsymbol{x}|t)} 
        + \beta(\alpha-v)\partial_v\rho(\boldsymbol{x}|t) \\
        - \beta\rho(\boldsymbol{x}|t)  \\
        - \frac{1}{2}\Big( v\partial_{y}^{2}\rho(\boldsymbol{x}|t) 
        + \sigma^2 v\partial_{v}^{2}\rho(\boldsymbol{x}|t) 
        + 2\rho\sigma v\partial_{yv}^{2}\rho(\boldsymbol{x}|t) \Big) \\
        - \rho\sigma \partial_{y}\rho(\boldsymbol{x}|t) 
        - \sigma^{2}\partial_{v}\rho(\boldsymbol{x}|t) = 0 
        & \text{$(v,y)\in \mathbb{R}^+\times \mathbb{R},\:t\in(0,T]$}, \\[6pt]
        \rho(v=0,y|t,0,\boldsymbol{x_0},\boldsymbol{\mu}) = 0 
        & \text{for $y\in \mathbb{R},\:t\in(0,T]$}, \\
        \rho(\boldsymbol{x}|0,0,\boldsymbol{x_0},\boldsymbol{\mu}) = \delta_{x_0}(\boldsymbol{x}) 
        & \text{for $(v,y)\in \mathbb{R}^+\times \mathbb{R},\:t\in\{0\}$}.
    \end{array}
    \end{cases}
\end{equation}
We adhere to the strict Feller condition \cite{Feller_condition} regarding the SDE parameters, specifically $\sigma<\sqrt{2\alpha\beta}$. This condition justifies the boundary condition $\rho(0,y|t,0,\boldsymbol{x_0},\boldsymbol{\mu})=0$.

\subsection{Structure of the Normalizing Flow}\label{sect:structure_flow_Heston}

The TPDF of \eqref{eq:ngnf_inf_Heston_model} can be decomposed as 
\begin{equation} \label{Decompose_Heston}
    p_{V,Y}(v,y|v_0,y_0) = p_{V}(v|v_0,y_0)p_{Y|V}(y|v,y_0,v_0) .
\end{equation}
This decomposition \eqref{Decompose_Heston} guides the construction of the transport map of the Flow $\boldsymbol{n}_{\theta}(v,y)$, which is defined by conditioning. The first component of $\boldsymbol{n}_{\theta}(v,y)$ updates the volatility variable, while the second component modifies the asset price conditioned on the volatility. 

We solve the equation \eqref{FP_Heston} in the bounded domain $\hat{\Omega}=(0,3)\times(-6.5,6.5)$, therefore, we set $\boldsymbol{\hat{a}}=[0,-6.5]$ and $\boldsymbol{\hat{b}}=[3,6.5]$.
The reference distribution of the Normalizing Flow is defined as the product of a Gamma distribution,
truncated over the domain $[0, 1]$, for the first component of the transport map (corresponding to the
volatility), and a uniform distribution over $[0, 1]$ for the second component (corresponding to the
asset price). The shape parameter of the Gamma distribution is set to 5/2, and its scale parameter
is set to 1/2. 
To work with reference distributions supported in the interval $[0,1]$, a linear scaling is applied after the last transformation \ref{eq:ngnf_inf_full_transformation_layer_l}. This scaling has the following expression
\begin{equation}\label{eq:ngnf_inf_scaling_last_layer}
    f(\boldsymbol{x})=\frac{\boldsymbol{x}+1}{2}
\end{equation}
and maps $[-1,1]^d$ to $[0,1]^d$. 
The truncated Gamma distribution allows us to impose the boundary condition in equation \eqref{FP_Heston}. 
Similar observations apply to the other numerical experiments conducted.
We use $L=4+1$ layers for the Normalizing Flow. The number of elements in each MTG is set to 7. The network contains two GRU layers, each with hidden size 8. This results in a parameter vector $\boldsymbol{\theta}(\tau)\in\mathbb{R}^{M}$, with $M=8448$. To approximate the Neural Galerkin residual \eqref{eq:ngnf_inf_J_cost_functional_Neural_Galerkin},
we choose $\eta(\boldsymbol{x}_0)$ as 
\begin{equation}\label{eq:ngnf_inf_prior_mu_x0}
\eta(\boldsymbol{x}_0)
=
\mathcal{U}(v_0;[0,0.25])\mathcal{U}(y_0;[-3.5,2.5]).
\end{equation}

We utilize a flat prior restricted to the parameters' domain for the Bayesian inference, namely
\begin{equation} \label{prior_heston}
    {\pi}_0(\boldsymbol{\mu})= \mathrm{1}_{\{\alpha>0\}}
    \mathrm{1}_{\{\beta>0\}}\mathrm{1}_{\{\sigma\in(0,\sqrt{2\alpha\beta})\}}\mathrm{1}_{\{-1<\rho< 1\}}\mathrm{1}_{\{\mu\in\mathbb{R}\}}.
\end{equation}
Notice that this distribution guarantees that the Feller condition is always satisfied. 

We train the model by sampling the parameters $\boldsymbol{\mu}$ from the following distribution 
\begin{equation}\label{prior_heston_train}
\begin{aligned}
\tilde{\pi}_0(\boldsymbol{\mu}) ={}& 
\mathrm{1}_{\{\alpha>0\}}\phi(\alpha|0.1,0.08)
\mathrm{1}_{\{\beta>0\}}\phi(\beta|3,0.2)\,\times \\
& \mathrm{1}_{\{\sigma\in(0,\sqrt{2\alpha\beta}]\}}\phi(\sigma|0.25,0.08)
\phi(\mu|0.05,0.03)\mathrm{1}_{\{-1\leq\rho\leq 1\}}\phi(\rho|-0.8,0.08).
\end{aligned}
\end{equation}
We employ a Monte-Carlo estimator based on $15000$ samples.
The dynamics satisfied by $\boldsymbol{\dot{\theta}}(\tau)$ \eqref{eq:ngnf_inf_optimization_Neural_Galerkin}, is integrated using an Explicit Runge-Kutta (RK) method of order 5(4), see Algorithm \ref{alg:ngnf_inf_integrate_NF_map}. The offline time required to integrate this dynamics (training time) was approximately 48h.\footnote{All experiments were implemented in PyTorch \cite{paszke2019pytorch} and run on a GPU cluster equipped with NVIDIA H100 SXM5 GPUs (4 GPUs per node, 94\,GB GPU memory per GPU).}

In Tables \ref{tab:heston_quantitative_error}, \ref{tab:heston_compare_quantitative_error_models}, we test the model on the distribution 
\begin{equation}\label{test_heston_distribution}
\begin{aligned}
\hat{\pi}(\boldsymbol{\mu}) ={}& 
\mathrm{1}_{\{\alpha>0\}}\phi(\alpha|0.075,0.1)
\mathrm{1}_{\{\beta>0\}}\phi(\beta|3,0.2)\,\times \\
& \mathrm{1}_{\{\sigma\in(0,\sqrt{2\alpha\beta}]\}}\phi(\sigma|0.25,0.1)
\phi(\mu|0.05,0.1)\mathrm{1}_{\{-1\leq\rho\leq 1\}}\phi(\rho|-0.8,0.1).
\end{aligned}
\end{equation}

In the numerical experiments, we run an MCMC chain of length 13000 and discard the first 3000 samples as burn-in.
For the implementation of MCMC in the form of slice sampling, we rely on the code released in \cite{pmlr-v89-papamakarios19a}.
The computational setup is summarized in Table \ref{tab:heston_nn_parameters}.

\begin{table}[!htb]
\centering
\begin{tabular}{lll}
\hline
\textbf{Quantity} & \textbf{Symbol} & \textbf{Value} \\
\hline
Diffusion domain & $\Omega$ & $\mathbb{R}^+\times\mathbb{R}$ \\
Parameter of the support of the NF (Volatility) & $[\hat{a}_1,\hat{b}_1]$ & [0,3] \\
Parameter of the support of the NF (log-Asset) & $[\hat{a}_2,\hat{b}_2]$ & [-6.5,6.5] \\
Initial condition distribution & $\eta(v_0,y_0)$ & See \eqref{eq:ngnf_inf_prior_mu_x0} \\
Parameter training distribution & $\tilde{\pi}_0(\boldsymbol{\mu})$ & See \eqref{prior_heston_train} \\
Parameter test distribution & $\hat{\pi}(\boldsymbol{\mu})$ & See \eqref{test_heston_distribution} \\
Distribution to approximate $J$ \eqref{eq:ngnf_inf_J_cost_functional_Neural_Galerkin} & $\nu_{{\theta}}(\cdot|\cdot,\cdot)$ & $\Rho(\cdot|\boldsymbol{\theta},\cdot,\cdot)$ \\
Number of layers & $L$ & $5$ \\
MTG elements per layer & -- & 7 \\
Number of GRU layers & -- & $2$ \\
Hidden size of each GRU layer & -- & $8$ \\
Dimension of $\boldsymbol{\theta}$ & $M$ & $8448$ \\
Number of Monte Carlo samples to approximate \eqref{eq:ngnf_inf_J_cost_functional_Neural_Galerkin} & $-$ & $15000$ \\
Integrator of the dynamics $\boldsymbol{\dot{\theta}}$ \eqref{eq:ngnf_inf_optimization_Neural_Galerkin} & -- & RK 5(4) \\
Offline GPU Training time & -- & $\simeq 48\mathrm{h}$ \\ 
Burn-in length & -- & 3000 \\
Length of the Markov chain & -- & 10000 \\
\hline
\end{tabular}
\caption{Computational setup, Normalizing Flow architecture, and parameter-sampling details for the Heston model \eqref{eq:ngnf_inf_Heston_model}.}
\label{tab:heston_nn_parameters}
\end{table}

\begin{algorithm}[t]
      \begin{flushleft}
        \textbf{INPUT:} Normalizing Flow $\boldsymbol{n}_{\theta}$, reference distribution $Z$, prior over SDE parameters $\tilde{\pi}_0^{\star}$, density for the location of the initial condition $\eta(\boldsymbol{x_0})$, integration time-horizon $\Delta$ \\
        \textbf{OUTPUT:} Set of parameters $\{\boldsymbol{\theta}(\tau)\}_{\tau\in I}$
    \end{flushleft}
    \begin{algorithmic}[1]

      \State $\boldsymbol{\theta}(0)$ 
      \Comment{Initialize weights and biases of the NF (cf. section \ref{sect:ngnf_inf_dirac_delta_imposition_zero_flux}).}
      \State $\tau=0$

      \While {$\tau<\Delta$}
        \State $\{\boldsymbol{x}_{0,i}\}_{i=1}^{N}\sim\eta$
        \Comment{Draw a batch of locations of the Dirac delta initial conditions.}
        \State $\{\boldsymbol{\mu}_{i}\}_{i=1}^{N}\sim\tilde{\pi}_0^{\star}$
         \Comment{Draw a batch of parameters.}

        \State $\{\boldsymbol{x}_{i}\}_{i=1}^{N}\sim\{\Rho(\cdot|\boldsymbol{\theta}(\tau),\boldsymbol{x}_{0,i},\boldsymbol{\mu}_{i})\}_{i=1}^N$
        \Comment{Use Algorithm \ref{approximated_sample_from_NF} to sample from the NF.}
        \State $\mathrm{f}_{\tau,i}=\mathcal{L}^{\star}_{s_i+\tau}(\Rho(\cdot|\boldsymbol{\theta}(\tau),\boldsymbol{x}_{0,i},\boldsymbol{\mu}_{i}))(\boldsymbol{x}_{i})$
        \State $(\mathrm{J})_{ik} = \frac{\partial \Rho(\boldsymbol{x}_{i}|\boldsymbol{\theta}(\tau),\boldsymbol{x}_{0,i},\boldsymbol{\mu}_{i})}{\partial \theta_k}$
        
        \State ${\boldsymbol{\dot{\theta}}}(\tau) \in \argmin_{\boldsymbol{\zeta}\in\dot{\Theta}} J_{\tau}(\boldsymbol{\theta}(\tau),\boldsymbol{\zeta})$ 
        \Comment{Using LSMR, solve \eqref{eq:ngnf_inf_J_cost_functional_Neural_Galerkin}.}

        \State Update $\boldsymbol{\theta}(\tau)$ from ${\boldsymbol{\dot{\theta}}}(\tau)$ according to the RK5(4) integrator.
        \State Update $\tau$ according to the RK5(4) integrator.
        
      \EndWhile
 
      \end{algorithmic}
  
      \caption{     
      Integration of the Neural Galerkin dynamics of the parameters of the Normalizing Flow \eqref{eq:ngnf_inf_optimization_Neural_Galerkin}.}
      \label{alg:ngnf_inf_integrate_NF_map}
    \end{algorithm}

\subsubsection{Additional Numerical Experiments for the Heston Model}\label{sect:appendix_numerical_experiments}

In Table~\ref{tab:heston_compare_quantitative_error_models}, we report the relative error between the reference log-likelihood and the approximated log-likelihood, both at the true parameter $\boldsymbol{\mu}^{\star}$ and averaged over a batch $\{\boldsymbol{\mu}^{i}\}_{i=1}^{N_{\mathrm{test}}}\sim\hat{\pi}$ (cf. \eqref{test_heston_distribution}) with $N_{\mathrm{test}}=100$, 
for NF and for the 2nd order closed-form expansions of Aït-Sahalia (A.-S.) \cite{ait_sahalia2008closed} and the 4th order It\^o-Taylor expansion \cite{YANG2019256}\footnote{Code for Aït-Sahalia's expansion is publicly available at
\url{https://www.princeton.edu/~yacine/research.htm}.
At the time of our experiments, the implementation of \cite{YANG2019256}
was available from the authors upon request.}. 
The Fourier inversion of the characteristic function of the Heston model gives the reference. Among the benchmarked approximations, NF attains the smallest trajectory log-likelihood errors.
\begin{table}[H]
\centering
\caption{Accuracy of the likelihood approximations for the Heston model \eqref{eq:ngnf_inf_Heston_model}.  
Best values in bold. The data are acquired with $\Delta=0.5$.}
\label{tab:heston_compare_quantitative_error_models}
\begin{tabular}{lccc}
\toprule
Metric & NF & A.-S. & It\^o-Taylor \\
\midrule
Relative error over the log-lh \eqref{eq:ngnf_inf_Factor_likelihood} given $\boldsymbol{\mu}^{\star}$ 
& \textbf{0.0003} & 0.0064 & 0.0093 \\
\midrule
Mean err. relative log-lh \eqref{eq:ngnf_inf_Factor_likelihood} given $\{\boldsymbol{\mu}^{i}\}_{i=1}^{N_{\mathrm{test}}}$ 
& \textbf{0.063}  & 0.401 & 0.361 \\
Standard err. (mean err. relative log-lh)
& \textbf{0.012}  & 0.078 & 0.089 \\
Median err. relative log-lh \eqref{eq:ngnf_inf_Factor_likelihood} given $\{\boldsymbol{\mu}^{i}\}_{i=1}^{N_{\mathrm{test}}}$ 
& \textbf{0.013}  & 0.193 & 0.064 \\
\bottomrule
\end{tabular}
\end{table}

\begin{figure}[!htb]
\centering

\begin{minipage}[t]{0.27\textwidth}
    \centering
    \includegraphics[width=\linewidth]{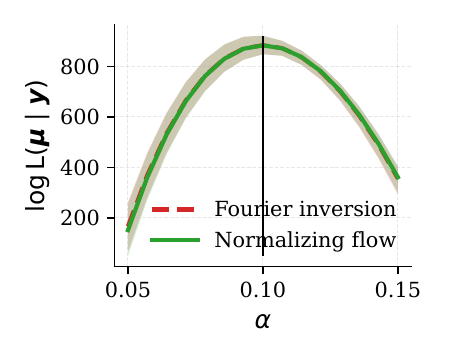}
\end{minipage}\hfill
\begin{minipage}[t]{0.27\textwidth}
    \centering
    \includegraphics[width=\linewidth]{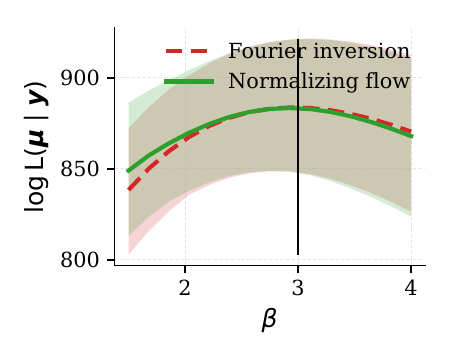}
\end{minipage}\hfill
\begin{minipage}[t]{0.27\textwidth}
    \centering
    \includegraphics[width=\linewidth]{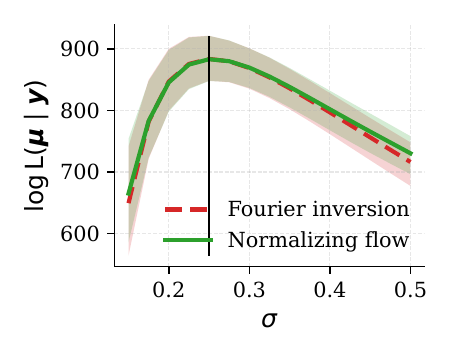}
\end{minipage}

\vspace{0.8em}

\makebox[\textwidth][c]{%
    \begin{minipage}[t]{0.27\textwidth}
        \centering
        \includegraphics[width=\linewidth]{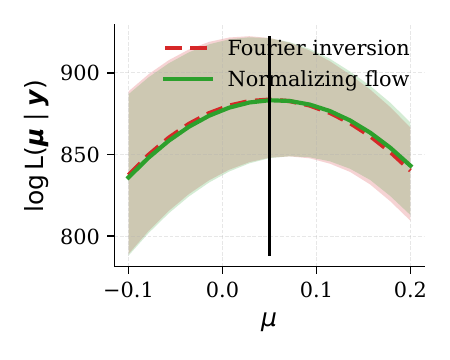}
    \end{minipage}\hspace{0.04\textwidth}
    \begin{minipage}[t]{0.27\textwidth}
        \centering
        \includegraphics[width=\linewidth]{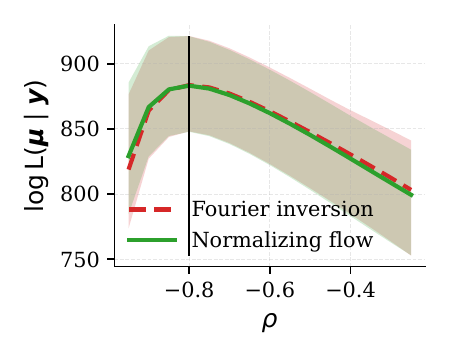}
    \end{minipage}
}

\caption{Log-likelihood profiles for the Heston model \eqref{eq:ngnf_inf_Heston_model} with respect to each parameter, while fixing the others at their corresponding values of $\boldsymbol{\mu}^{\star}$. The red dashed curves correspond to the semi-closed-form log-likelihood, and the green curves to the Normalizing Flow approximation. The black vertical line marks the true parameter value. The curves are computed over 25 independently generated datasets, and the shaded bands represent pointwise 95\% confidence intervals across datasets. The sampling frequency is set to $\Delta=0.5$.}
\label{fig:ngnf_inference_NG_vs_Fourier_log_lh_heston_sensitivity}
\end{figure}

Table \ref{tab:runtime_comparison_heston_various_methods} presents a comparison of the performance of the approximated TPDF for sampling from the posterior distribution. NF achieves the shortest time per effective sample and the smallest empirical Wasserstein $2$ distance in relation to the Fourier reference.
\begin{table}[!htb]
\centering
\caption{Heston model \eqref{eq:ngnf_inf_Heston_model}. Runtime for $10^4$ MCMC samples, empirical joint $W_2$ distance, and marginal parameter-wise $W_2$ distances relative to the Fourier reference. Best values in bold. A.-S. is omitted as it does not target the posterior.}
\label{tab:runtime_comparison_heston_various_methods}
\begin{tabular}{llccccccc}
\toprule
Model & Method & Time (s) & $W_2$ & $W_2^\alpha$ & $W_2^\beta$ & $W_2^\sigma$ & $W_2^\mu$ & $W_2^\rho$ \\
\midrule
\multirow{3}{*}{Heston}
& Fourier & 14560 & Ref. & -- & -- & -- & -- & -- \\
& NF & $\boldsymbol{2243}$ & $\boldsymbol{0.066}$ & $\boldsymbol{0.0003}$ & $\boldsymbol{0.057}$ & $\boldsymbol{0.004}$ & $\boldsymbol{0.0046}$ & $\boldsymbol{0.0110}$ \\
& It\^o-Taylor & 15069 & 0.948 & 0.0071 & 0.938 & 0.0918 & 0.0518 & 0.066 \\
\bottomrule
\end{tabular}
\end{table}
To demonstrate the robustness of the likelihood profiles with respect to the simulated trajectory, we repeated the likelihood analysis on 25 independent datasets generated using the same $\boldsymbol{\mu}^{\star}$, with all datasets supported on $\eta(\boldsymbol{x}_0)$, i.e., contained in the support of the measure $\eta$. The resulting log-likelihood slices, along with pointwise confidence bands, are presented in Figure \ref{fig:ngnf_inference_NG_vs_Fourier_log_lh_heston_sensitivity}.

\begin{figure}[!htb]
\centering

\begin{minipage}[t]{0.27\textwidth}
    \centering
    \includegraphics[width=\linewidth]{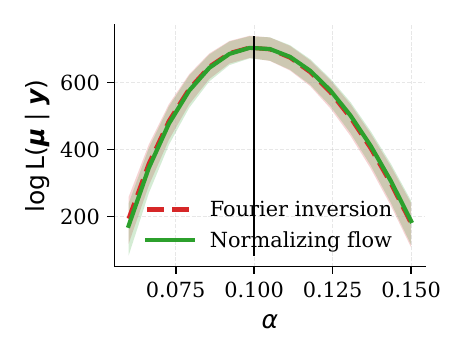}
\end{minipage}\hfill
\begin{minipage}[t]{0.27\textwidth}
    \centering
    \includegraphics[width=\linewidth]{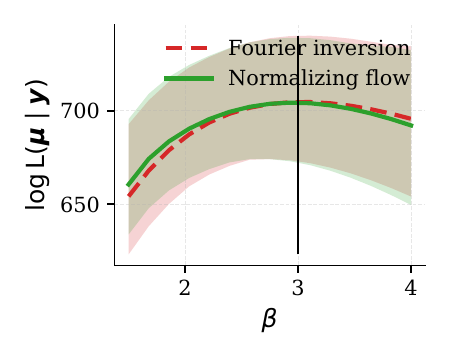}
\end{minipage}\hfill
\begin{minipage}[t]{0.27\textwidth}
    \centering
    \includegraphics[width=\linewidth]{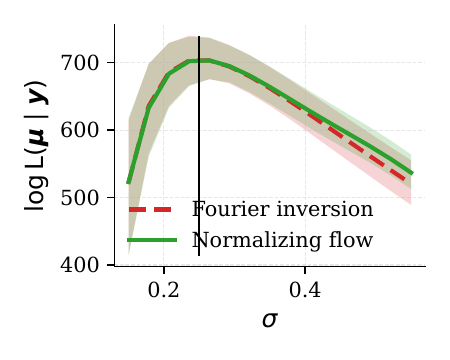}
\end{minipage}

\vspace{0.8em}

\makebox[\textwidth][c]{%
    \begin{minipage}[t]{0.27\textwidth}
        \centering
        \includegraphics[width=\linewidth]{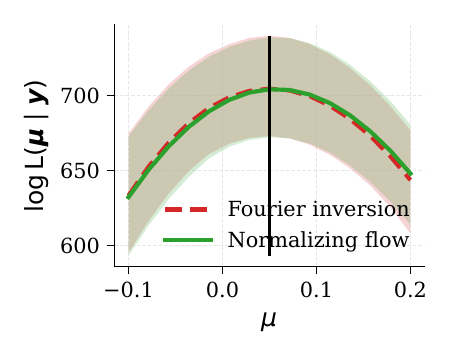}
    \end{minipage}\hspace{0.04\textwidth}
    \begin{minipage}[t]{0.27\textwidth}
        \centering
        \includegraphics[width=\linewidth]{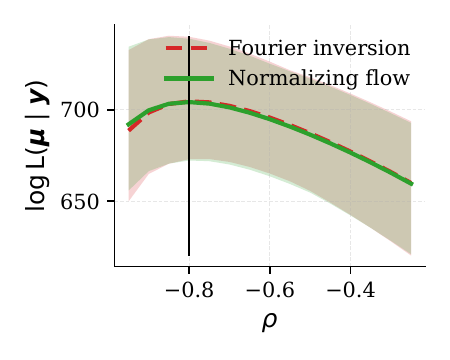}
    \end{minipage}
}

\caption{Log-likelihood profiles for the Heston model \eqref{eq:ngnf_inf_Heston_model}, obtained by varying one component of $\boldsymbol{\mu}^{\star}$ at a time while keeping the others fixed. The red dashed curves correspond to the semi-closed-form log-likelihood, and the green curves to the Normalizing Flow approximation. The black vertical line marks the true parameter value. The curves are computed over 25 independently generated datasets and the shaded bands represent pointwise 95\% confidence intervals across datasets. The sampling frequency is set to $\Delta=1$.}
\label{fig:ngnf_inference_NG_vs_Fourier_log_lh_heston_sensitivity_delta_1}
\end{figure}

We further evaluate the robustness of our model by using data sampled at a lower frequency, specifically with \(\Delta = 1\) and $n+1=200$.  We integrate the dynamics \eqref{eq:ngnf_inf_optimization_Neural_Galerkin} over the time interval $[0,\Delta]$ and conduct the same sensitivity analysis as shown in Figure \ref{fig:ngnf_inference_NG_vs_Fourier_log_lh_heston_sensitivity}. To achieve this, as done above, we generate 25 independent datasets and compute the likelihood profiles for each. The true parameter $\boldsymbol{\mu}^{\star}$ is fixed as in the previous experiments. The results are shown in Figure \ref{fig:ngnf_inference_NG_vs_Fourier_log_lh_heston_sensitivity_delta_1}.

Our findings indicate that, in both observation regimes, our model qualitatively matches the width and shape of the confidence intervals produced by the semi-closed-form solution. This suggests that the Normalizing Flow is robust across various observation settings, supporting its use for Bayesian inference of diffusion processes.

We conduct a final test to evaluate the robustness of the model. Specifically, we generate 25 independent datasets $\{\boldsymbol{y}^i\}_{i=1}^{25}$ with $\Delta=0.5$, the same parameter $\boldsymbol{\mu}^{\star}$ and, for each of them, we run MCMC in the form of slice sampling to approximate the posterior distribution. We then compute the corresponding posterior means, and summarize these results across datasets. The comparison between the Fourier-based reference posterior and the Normalizing Flow approximation is presented in Table~\ref{tab:heston_post_mean_across_datasets}.  Our results indicate that the posterior summaries obtained from Fourier inversion and the Normalizing Flow closely align for all parameters.
\begin{table}[t]
\caption{Heston model \eqref{eq:ngnf_inf_Heston_model} with $\Delta =0.5$; posterior means averaged across datasets, with pointwise 95\% confidence intervals across datasets.}
\label{tab:heston_post_mean_across_datasets}
\centering
\begin{tabular}{lccc}
\toprule
Parameter & True parameter $\boldsymbol{\mu}^{\star}$ & Fourier inversion & Normalizing Flow \\
\midrule
$\alpha$ & 0.10 & 0.1003 [0.0967, 0.1028] & 0.1000 [0.0971, 0.1024] \\
$\beta$  & 3.00 & 3.087 [2.479, 3.596] & 3.1279 [2.5684, 3.6564] \\
$\sigma$ & 0.25 & 0.2549 [0.2319, 0.2764] & 0.2601 [0.2380, 0.2839] \\
$\mu$    & 0.05 & 0.0479 [0.0229, 0.0761] & 0.0565 [0.0303, 0.0859] \\
$\rho$   & -0.80 & -0.7977 [-0.8418, -0.7548] & -0.8135 [-0.8637, -0.7657] \\
\bottomrule
\end{tabular}
\end{table}

\subsubsection{MCMC diagnostic for the Heston model}

\begin{figure}[h!]
\minipage{0.3\textwidth}
  \includegraphics[width=\linewidth]{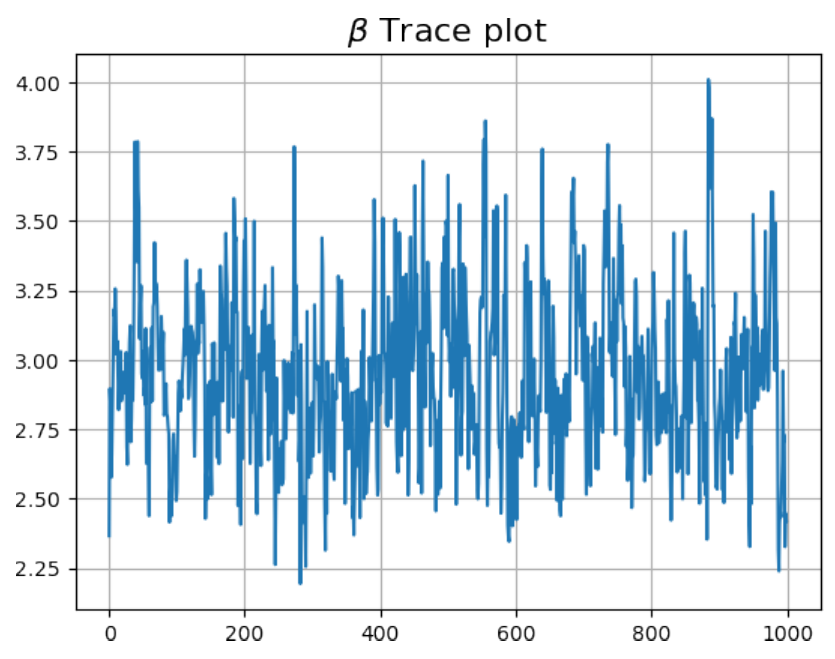}
\endminipage\hfill
\minipage{0.3\textwidth}
  \includegraphics[width=\linewidth]{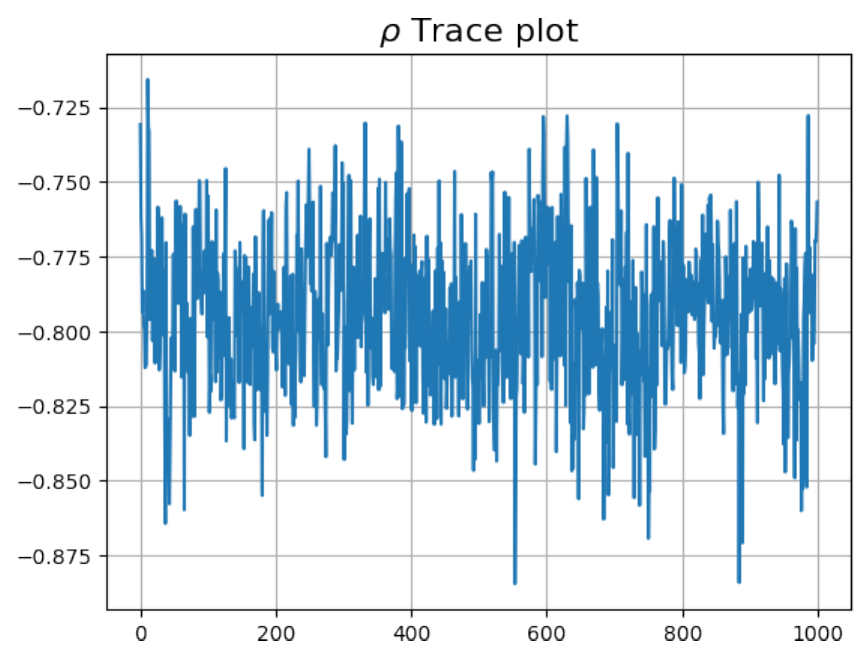}
\endminipage\hfill
\minipage{0.3\textwidth}
  \includegraphics[width=\linewidth]{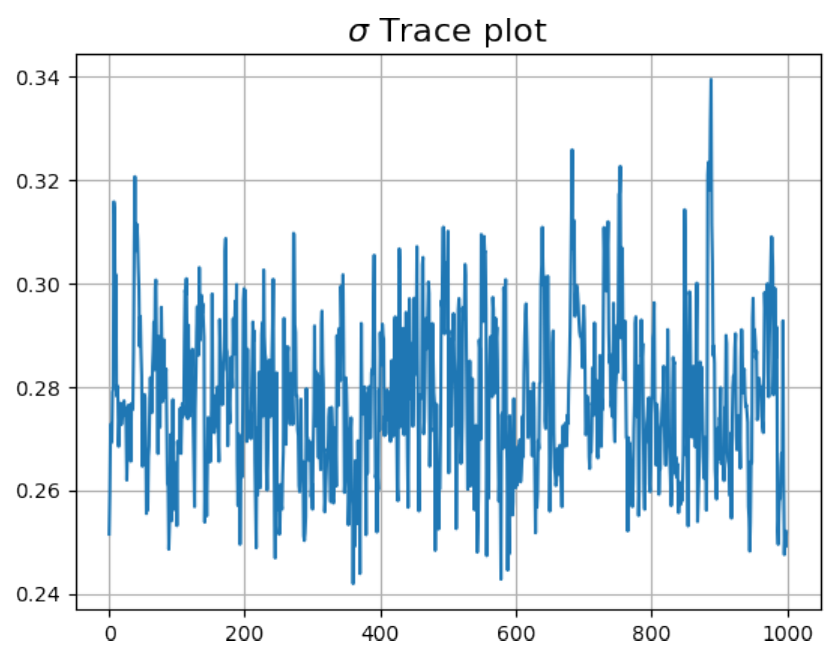}
\endminipage\hfill 

\minipage{0.3\textwidth}
  \includegraphics[width=\linewidth]{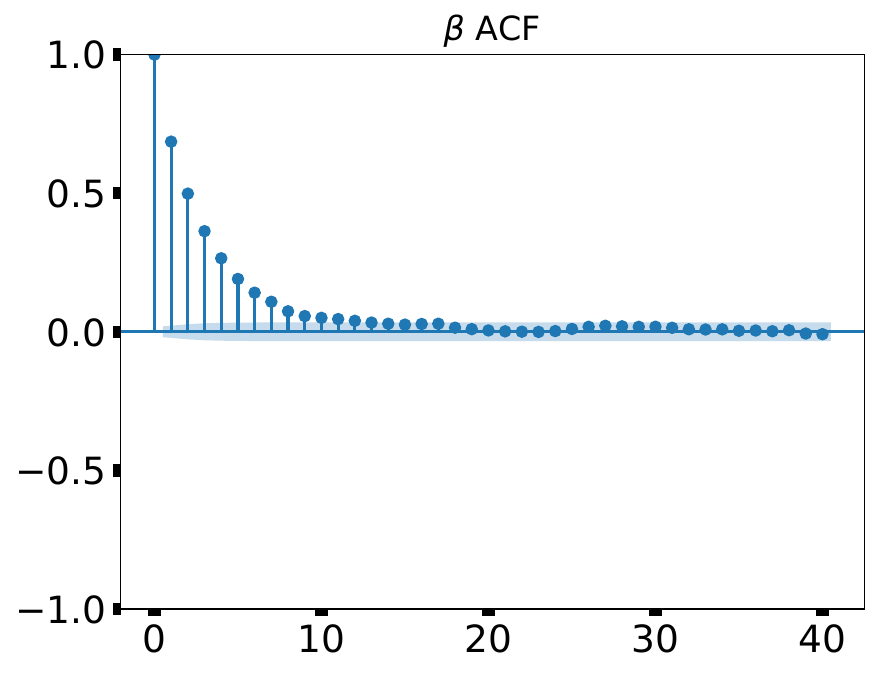}
\endminipage\hfill
\minipage{0.3\textwidth}
  \includegraphics[width=\linewidth]{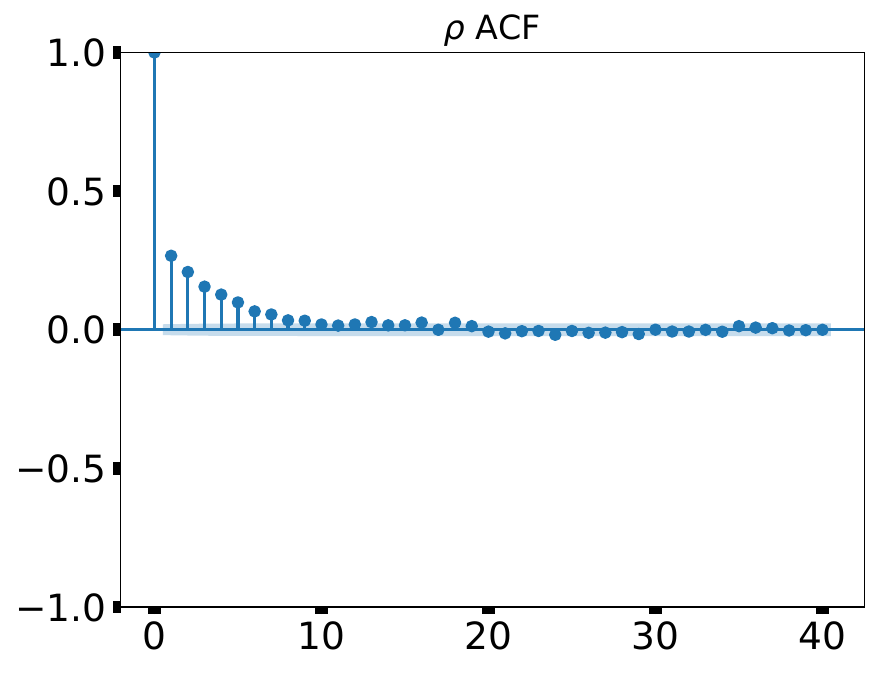}
\endminipage\hfill
\minipage{0.3\textwidth}
  \includegraphics[width=\linewidth]{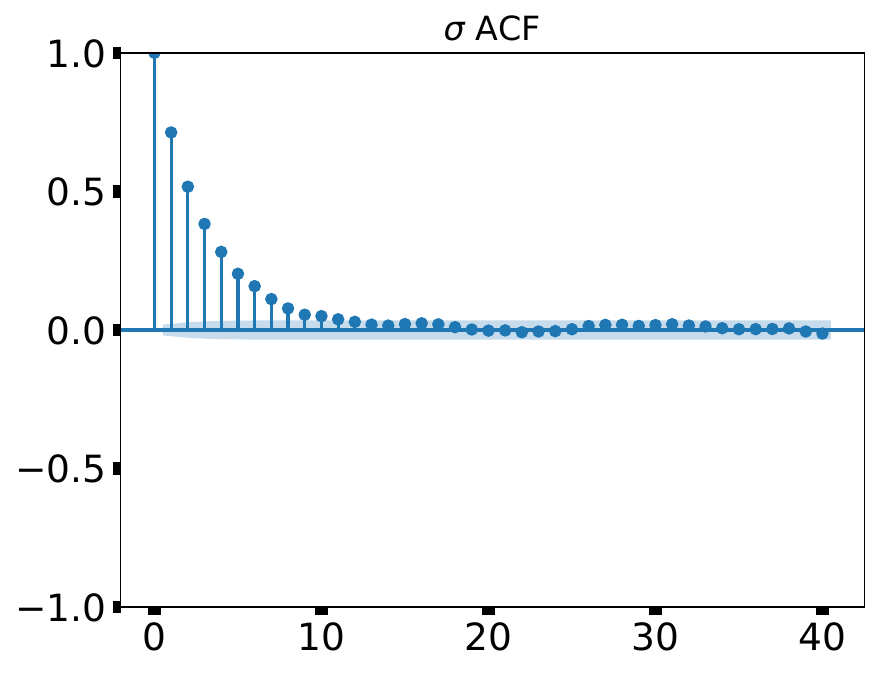}
\endminipage\hfill 

\caption{Trace plots and autocorrelation function for the parameters $\beta$, $\rho$ and $\sigma$ of the Heston model \eqref{eq:ngnf_inf_Heston_model}. The chain, drawn with MCMC in the form of axis-aligned, is composed of 10000 samples.}
\label{fig:ngnf_inf_ACF_Heston}
\end{figure}

We diagnose the performance of the Flow-based MCMC sampler by investigating the chain's mixing time. This indicator, as shown in Figure \ref{fig:ngnf_inf_ACF_Heston}, is excellent.

\subsubsection{Additional Numerical Illustrations}\label{sect:additional_illustrations_heston}

\begin{figure}[!htb]
\centering

\begin{minipage}[t]{0.27\textwidth}
    \centering
    \includegraphics[width=\linewidth]{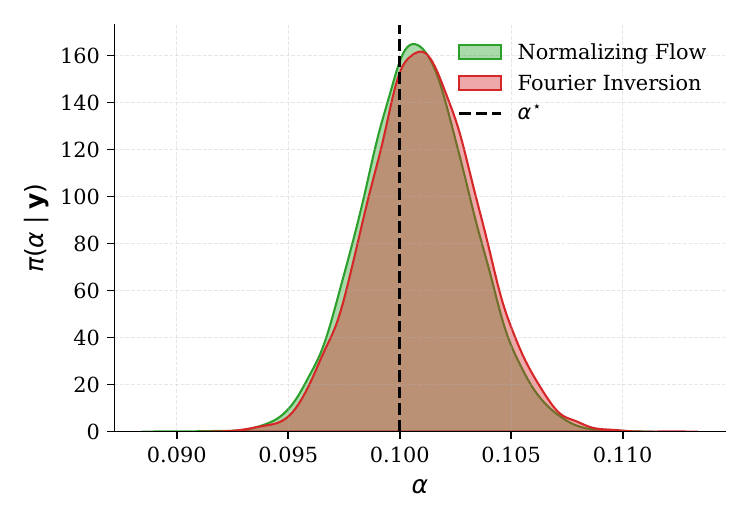}
\end{minipage}\hfill
\begin{minipage}[t]{0.27\textwidth}
    \centering
    \includegraphics[width=\linewidth]{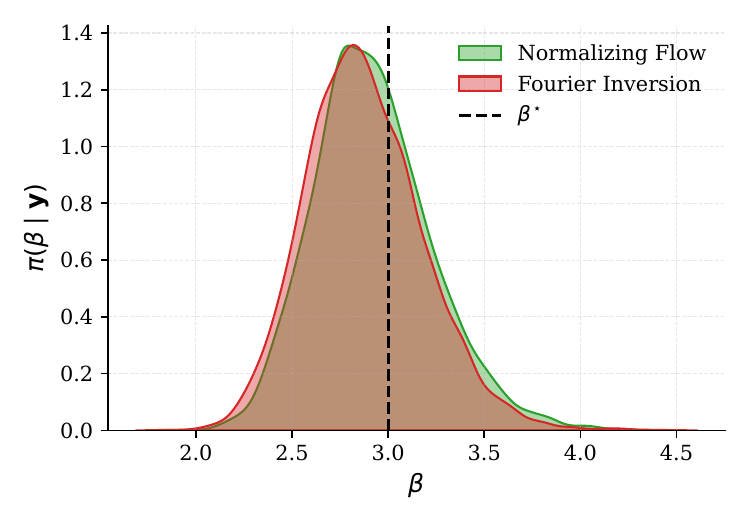}
\end{minipage}\hfill
\begin{minipage}[t]{0.27\textwidth}
    \centering
    \includegraphics[width=\linewidth]{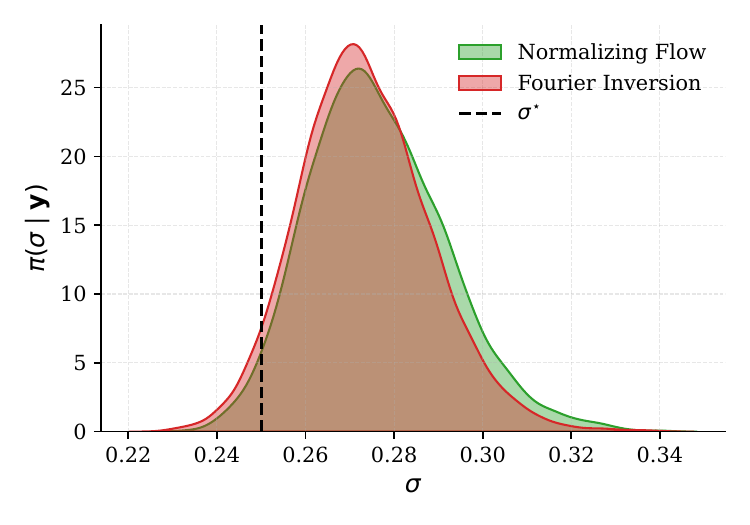}
\end{minipage}

\vspace{0.8em}

\makebox[\textwidth][c]{%
    \begin{minipage}[t]{0.27\textwidth}
        \centering
        \includegraphics[width=\linewidth]{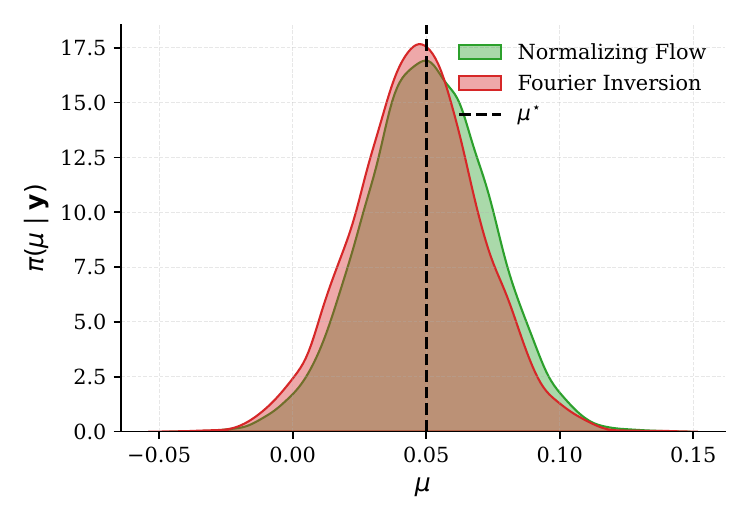}
    \end{minipage}\hspace{0.04\textwidth}
    \begin{minipage}[t]{0.27\textwidth}
        \centering
        \includegraphics[width=\linewidth]{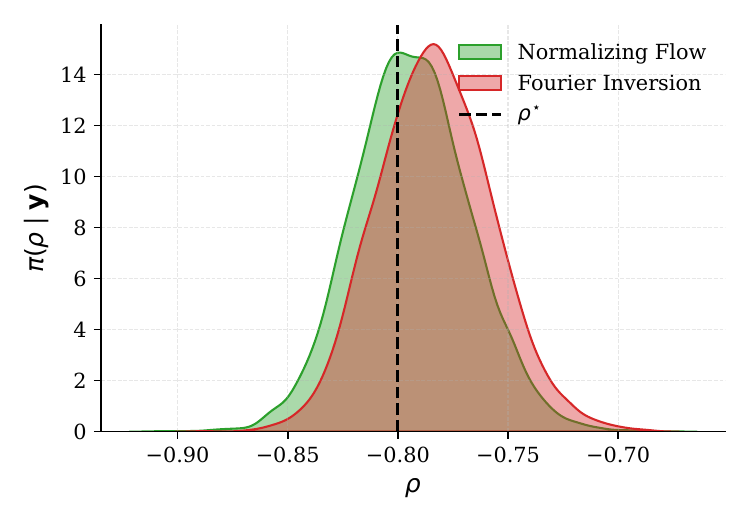}
    \end{minipage}
}

\caption{Posterior marginal distributions for the parameters of the Heston model \eqref{eq:ngnf_inf_Heston_model}. }
\label{fig:ngnf_inference_Posterior_heston_marginal}
\end{figure}

In Figure \ref{fig:ngnf_inference_Posterior_heston_marginal}, we report the marginal distribution of each parameter obtained by performing a Kernel Density plot on a chain of $10^4$ samples. As visible in Figure \ref{fig:ngnf_inf_NG_vs_Fourier_log_lh_heston}, our model slightly overestimates the likelihood slice concerning the parameter $\rho$, and this effect is reflected in the corresponding posterior marginal distribution. The other posterior distributions show a close agreement.

\begin{figure}[!htb]
\minipage{0.35\textwidth}
  \includegraphics[width=\linewidth]{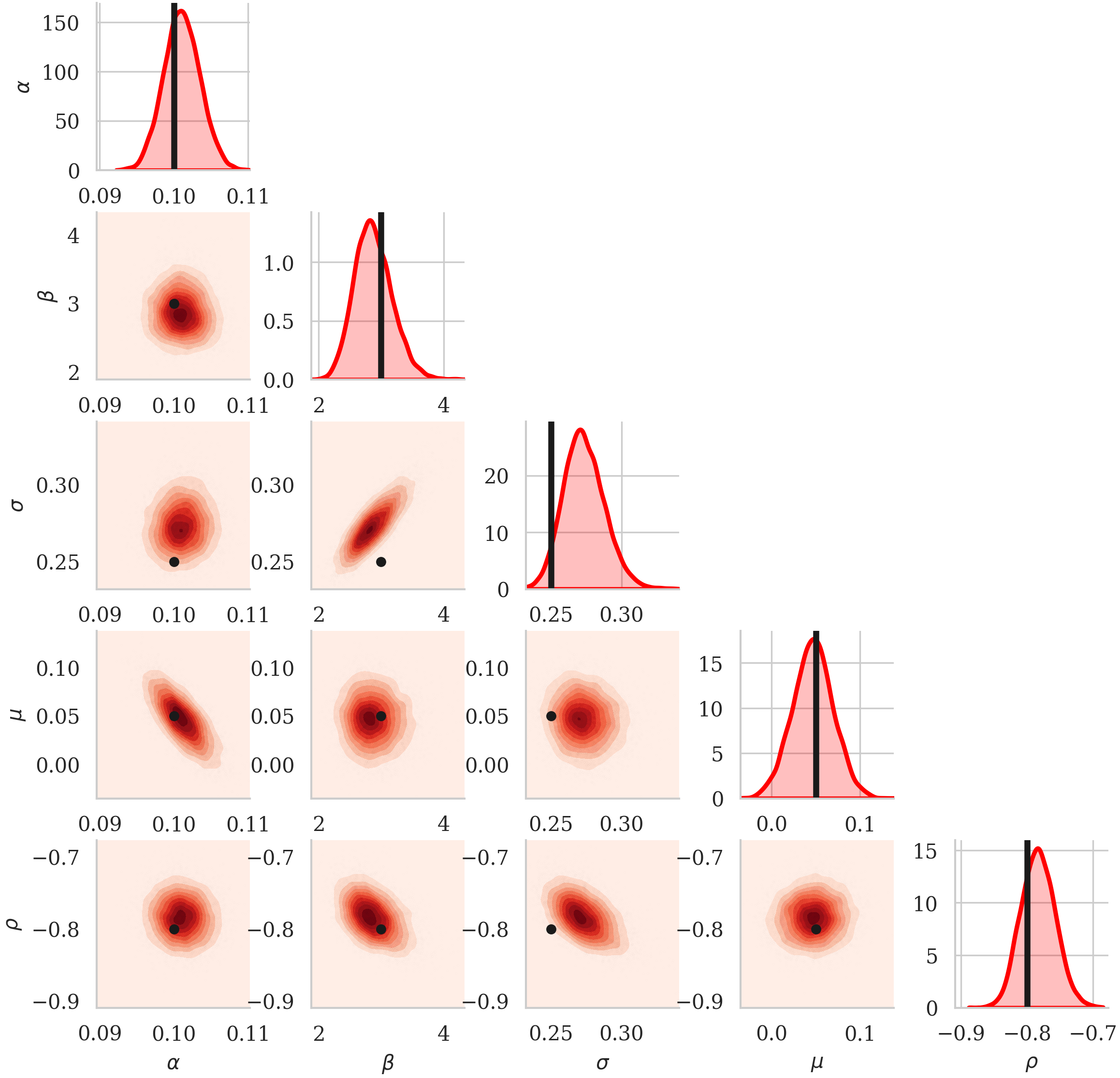}
\endminipage\hfill
\minipage{0.35\textwidth}
  \includegraphics[width=\linewidth]{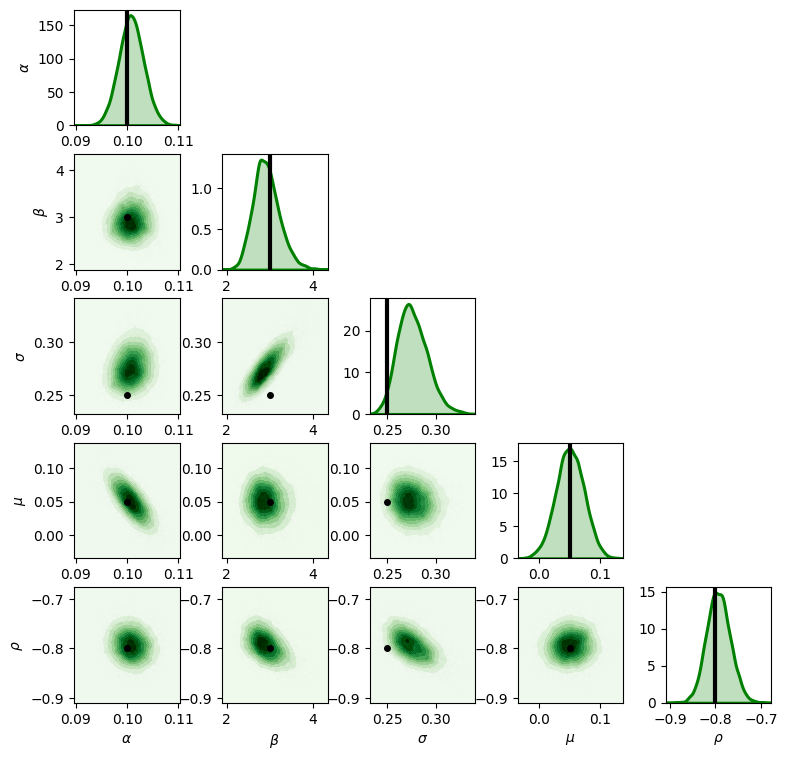}
\endminipage\hfill
\caption{Samples drawn from the posterior distribution of the Heston model. In the left figure, the TPDF is obtained by Fourier inversion. In the right figure, the TPDF is approximated by Normalizing Flows.}
\label{fig:ngnf_inf_NG_posterior_heston}
\end{figure}

Joint posterior plots for the Heston model, showing that the surrogate captures posterior correlations, are reported in Figure \ref{fig:ngnf_inf_NG_posterior_heston}.

\section{Setup for the SVCEV Stochastic Volatility Model}\label{sect:ngnf_inf_svcev_subsection_supplementary}

We describe the experimental setup for the SVCEV model.
Using the same simulation strategy employed for the Heston model, we utilize the Euler-Maruyama scheme to simulate a trajectory consisting of $n+1=100$ observations with parameters $\alpha^{\star}=0.1$, $\beta^{\star}=3$, $\sigma^{\star}=0.25$, $\mu^{\star}=0.05$, $\rho^{\star}=-0.8$, $\gamma^{\star}=1$, $\Delta=0.5$.

By conducting similar computations to those in \cite{Feller_condition}, it can be shown that for $\gamma\in[0.5,1]$ and $\sigma^2\leq 2\alpha\beta$, the origin is inaccessible. This translates into a zero Dirichlet boundary condition in the FP equation $\rho(v=0,y|t,0,\boldsymbol{x_0},\boldsymbol{\mu}) = 0 $.

Setting $\boldsymbol{x}=(v,y)$, the FP equation associated with \eqref{eq:ngnf_inf_SVCEV} reads
\begin{equation} \label{FP_SVCEV}
    \begin{array}{ll}
    \begin{cases}
        \partial_t{\rho(\boldsymbol{x}|t)} + (\mu - v/2)\partial_y{\rho(\boldsymbol{x}|t)} 
        + \beta(\alpha-v)\partial_v\rho(\boldsymbol{x}|t) - \beta\rho(\boldsymbol{x}|t) \\
        -\frac{1}{2}\Big( v\partial_{y}^{2}\rho(\boldsymbol{x}|t) 
        + \sigma^2 v^{2\gamma}\partial_{v}^{2}\rho(\boldsymbol{x}|t) 
        + 2\rho\sigma v^{\gamma+0.5}\partial_{yv}^{2}\rho(\boldsymbol{x}|t) \Big) \\
        - (\gamma+\frac{1}{2})\rho\sigma v^{\gamma-0.5}\partial_{y}\rho(\boldsymbol{x}|t) 
        - 2\gamma\sigma^{2}v^{2\gamma-1}\partial_{v}\rho(\boldsymbol{x}|t) \\
        - \frac{1}{2}(2\gamma (2\gamma-1)\sigma^2v^{2(\gamma-1)})\rho(\boldsymbol{x}|t) = 0 
        & \text{$(v,y)\in\mathbb{R}^+\times\mathbb{R},\:t\in(0,T]$}, \\[6pt]
        \rho(v=0,y|t,0,\boldsymbol{x_0},\boldsymbol{\mu}) = 0 
        & \text{for $y\in\mathbb{R},\:t\in(0,T]$}, \\
        \rho(\boldsymbol{x}|0,0,\boldsymbol{x_0},\boldsymbol{\mu}) = \delta_{x_0}(\boldsymbol{x}) 
        & \text{for $(v,y)\in\mathbb{R}^+\times\mathbb{R}, t\in\{0\}$}.
    \end{cases}
    \end{array}
\end{equation}
The determinant of the covariance matrix of the SVCEV model is given by $\det(\Sigma(t,v,y))=\sigma^2v^{2\gamma+1}(1-\rho^2)$. This expression makes the solution of \eqref{FP_SVCEV} more challenging than in the affine case where $\gamma=0.5$, as the determinant approaches zero at a rate that increases with $\gamma$. 

The TPDF is decomposed as done for the Heston model \eqref{Decompose_Heston}; the map $\boldsymbol{n}_{\theta}$ updates the volatility independently while conditioning the asset price on it.
To approximate the Neural Galerkin residual \eqref{eq:ngnf_inf_J_cost_functional_Neural_Galerkin},
we choose $\eta(\boldsymbol{x}_0)$ as 
\begin{equation}\label{eq:ngnf_inf_prior_mu_x0_SVCEV}
\eta(\boldsymbol{x}_0)
=
\mathcal{U}(v_0;[0,0.25])\mathcal{U}(y_0;[-4,3]).
\end{equation}

The distribution $\tilde{\pi}_0$ is chosen as 
\begin{equation} \label{prior_svcev_train}
    \begin{split}
    \tilde{\pi}_0(\boldsymbol{\mu})= & \mathrm{1}_{\{\alpha>0\}}\phi(\alpha|0.1,0.08)
    \mathrm{1}_{\{\beta>0\}}\phi(\beta|3,0.2)\times \\ 
    & \mathrm{1}_{\{\sigma\in(0,\sqrt{2\alpha\beta}]\}}\phi(\sigma|0.25,0.08)\phi(\mu|0.05,0.03)\mathrm{1}_{\{-1\leq\rho\leq 1\}}\phi(\rho|-0.8,0.08)\times \\ 
    & \mathrm{1}_{\{0.5\leq\gamma\leq 1\}}\phi(\gamma|1,0.2).
    \end{split}
\end{equation}
while the prior is set as 
\begin{equation} \label{prior_svcev}
    {\pi}_0(\boldsymbol{\mu})= \mathrm{1}_{\{\alpha>0\}}
    \mathrm{1}_{\{\beta>0\}}\mathrm{1}_{\{\sigma\in(0,\sqrt{2\alpha\beta})\}}\mathrm{1}_{\{-1\leq\rho\leq 1\}}\mathrm{1}_{\{0.5\leq\gamma\leq 1\}}.
\end{equation}
The reference distribution of the Normalizing Flow is set as for the Heston model (see section \ref{sect:ngnf_inf_Heston_subsection_supplementary}). 
The linear scaling defined in \eqref{eq:ngnf_inf_scaling_last_layer} is used to map the variables into $[0,1]^d$ prior to applying the reference distributions.

Instead of solving for \eqref{FP_SVCEV}, we found it more stable to solve the PDE for the log-density: $q(\boldsymbol{x}|\boldsymbol{\theta}(\tau),\boldsymbol{x_0},\boldsymbol{\mu})=\mathrm{log}(\Rho(\boldsymbol{x}|\boldsymbol{\theta}(\tau),\boldsymbol{x_0},\boldsymbol{\mu}))$. The increased stability we achieve is related to the fast decay of the initial condition, which, after the log-transformation, becomes more manageable.

At inference time, in Table \ref{tab:svcev_quantitative_error}, we test the model on the distribution
\begin{equation}\label{test_svcev_distribution}
\begin{aligned}
\hat{\pi}(\boldsymbol{\mu}) ={}& 
\mathrm{1}_{\{\alpha>0\}}\phi(\alpha|0.1,0.03)
\mathrm{1}_{\{\beta>0\}}\phi(\beta|3,0.2)\,\times \\
& \mathrm{1}_{\{\sigma\in(0,\sqrt{2\alpha\beta}]\}}\phi(\sigma|0.25,0.03)
\phi(\mu|0.05,0.03)\mathrm{1}_{\{-1\leq\rho\leq 1\}}\phi(\rho|-0.8,0.07)\times \\ & \mathrm{1}_{\{0.5\leq\gamma\leq 1\}}\phi(\gamma|1,0.2).
\end{aligned}
\end{equation}

The remaining computational parameters and training details are reported in Table 
\ref{tab:svcev_nn_parameters}.
\begin{table}[!htb]
\centering
\begin{tabular}{lll}
\hline
\textbf{Quantity} & \textbf{Symbol} & \textbf{Value} \\
\hline
Diffusion domain & $\Omega$ & $\mathbb{R}^+\times\mathbb{R}$ \\
Parameter of the support of the NF (Volatility) & $[\hat{a}_1,\hat{b}_1]$ & [0,3] \\
Parameter of the support of the NF (log-Asset) & $[\hat{a}_2,\hat{b}_2]$ & [-6.5,6.5] \\
Initial condition distribution & $\eta(v_0,y_0)$ & See \eqref{eq:ngnf_inf_prior_mu_x0_SVCEV} \\
Parameter training distribution & $\tilde{\pi}_0(\boldsymbol{\mu})$ & See \eqref{prior_svcev_train} \\
Parameter test distribution & $\hat{\pi}(\boldsymbol{\mu})$ & See \eqref{test_svcev_distribution} \\
Distribution to approximate $J$ \eqref{eq:ngnf_inf_J_cost_functional_Neural_Galerkin} & $\nu_{{\theta}}(\cdot|\cdot,\cdot)$ & $\Rho(\cdot|\boldsymbol{\theta},\cdot,\cdot)$ \\
Number of Neural Galerkin samples & $N$ & $8066$ \\
Number of layers & $L$ & $5$ \\
MTG elements per layer & -- & $4$  \\
Number of GRU layers & -- & $4$ \\
Hidden size of each GRU layer & -- & $5$ \\
Dimension of $\boldsymbol{\theta}$ & $M$ & $6756$ \\
Number of Monte Carlo samples to approximate \eqref{eq:ngnf_inf_J_cost_functional_Neural_Galerkin} & $-$ & $13512$ \\
Integrator of the dynamics $\boldsymbol{\dot{\theta}}$ \eqref{eq:ngnf_inf_optimization_Neural_Galerkin} & -- & RK 5(4) \\
Offline GPU Training time & -- & $\simeq 48\mathrm{h}$ \\ 
Burn-in length & -- & 3000 \\
Length of the Markov chain & -- & 10000 \\
\hline
\end{tabular}
\caption{Computational setup, Normalizing Flow architecture, and parameter-sampling details for the SVCEV model \eqref{eq:ngnf_inf_SVCEV}.}
\label{tab:svcev_nn_parameters}
\end{table}

\subsection{Validation of the reference benchmark for the SVCEV Model}\label{sect:ngnf_inf_additional_experiments_svcev}

 We validate the accuracy of the data augmentation strategy used to benchmark the SVCEV model. The core idea of this strategy is to insert $\hat{M}-1$ latent variables between two physical observations and then integrate them out to express the TPDF as follows:
\begin{equation} \label{eq:ngnf_inf_Data_augmentation}
    p(\boldsymbol{y}_{t_i}|\Delta,\boldsymbol{y}_{t_{i-1}},\boldsymbol{\mu})=\int \prod_{m=0}^{\hat{M}-1} p(\boldsymbol{u}_{m+1}|\Delta/\hat{M},\boldsymbol{u}_{m},\boldsymbol{\mu})d\boldsymbol{u}_{1},...,\boldsymbol{u}_{\hat{M}-1},
\end{equation}
where $\boldsymbol{u}_{0}=\boldsymbol{y}_{t_{i-1}}$ and $\boldsymbol{u}_{\hat{M}}=\boldsymbol{y}_{t_{i}}$.
The integral in \eqref{eq:ngnf_inf_Data_augmentation} is not computable in general. Still, since the observation frequency has been increased by a factor $\hat{M}$, each term $p(\boldsymbol{u}_{m+1}|\Delta/\hat{M},\boldsymbol{u}_{m},\boldsymbol{\mu})$ can be approximated using the Euler-Maruyama TPDF. The integral can then be evaluated using Importance Sampling. The choice of the importance measure has been discussed in \cite{Durham01072002}. We use the so-called Modified Brownian Bridge (MBB); see p. 305 of \cite{Durham01072002}. Each term of the likelihood function is evaluated by sampling $N_{MC}$ independent realizations of the Modified Brownian Bridge and by approximating \eqref{eq:ngnf_inf_Data_augmentation} by its Monte Carlo estimator 
\begin{equation} \label{eq:ngnf_inf_Data_augmentation_mc}
    p(\boldsymbol{y}_{t_i}|\Delta,\boldsymbol{y}_{t_{i-1}},\boldsymbol{\mu})\simeq \frac{1}{N_{MC}}\sum_{k=1}^{N_{MC}} \frac{\prod_{m=0}^{\hat{M}-1}\phi(\boldsymbol{u}_{k,m+1}|\Delta/\hat{M},\boldsymbol{u}_{k,m},\boldsymbol{\mu})}{\prod_{m=0}^{\hat{M}-2}\mathrm{q}(\boldsymbol{u}_{k,m+1}|\Delta/\hat{M},\boldsymbol{u}_{k,m},\boldsymbol{\mu})},
\end{equation}
where $\phi$ denotes the TPDF associated with the Euler-Maruyama scheme with timestep size $\Delta/\hat{M}$ and $\mathrm{q}$ denotes the importance measure. 

We estimate the likelihood \eqref{eq:ngnf_inf_Factor_likelihood} as 
\begin{equation} \label{eq:ngnf_inf_Factor_likelihood_MBB}
    \hat{\mathrm{L}}(\boldsymbol{\mu}|\boldsymbol{y})=\prod_{i=1}^{n} \frac{1}{N_{MC}}\sum_{k=1}^{N_{MC}} \frac{\prod_{m=0}^{\hat{M}-1}\phi(\boldsymbol{u}_{i,k,m+1}|\Delta/\hat{M},\boldsymbol{u}_{i,k,m},\boldsymbol{\mu})}{\prod_{m=0}^{\hat{M}-2}\mathrm{q}(\boldsymbol{u}_{i,k,m+1}|\Delta/\hat{M},\boldsymbol{u}_{i,k,m},\boldsymbol{\mu})}.
\end{equation}

To assess the accuracy of the data-augmentation likelihood estimator, we compare coarser resolutions $(\hat{M},N_{MC})$ against a reference estimate $\widehat{\ell}_{\mathrm{ref}}$ obtained using the same estimator at a finer discretization, namely $(\hat{M},N_{MC})=(1500,3000)$. Starting from this reference resolution, we then evaluate additional estimators generated by successively halving both $\hat{M}$ and $N_{MC}$, including \((\hat{M},N_{MC}) = (1000, 2500)\). Table~\ref{tab:da_validation} shows that the relative errors at the finest resolutions are on the order of $10^{-4}$, and that the estimators tend to converge toward the reference value.
 
In light of these results, to get a reference for the experiments in Table \ref{tab:svcev_quantitative_error}, we employ $\hat{M}=1000$ and $N_{MC}=2500$, leading to $\Delta/\hat{M} = 5\times10^{-4}$. 
Instead, to keep the cost of the MCMC experiment in section \ref{sect:ngnf_inf_svcev} affordable, we sample the reference chain with $M=200$ and $N=400$.
\begin{table}[H]
\centering
\begin{tabular}{ccccc}
\hline
$\hat{M}$ & $N_{MC}$ & $\widehat{\ell}(\hat{M},N_{MC})$ & Relative error & 95\% CI (rel. error) \\
\hline
1000 & 2500 & 368.979 & $5.46\times 10^{-4}$ & $[4.96,\,5.96]\times 10^{-4}$ \\
750  & 1500 & 368.972 & $7.17\times 10^{-4}$ & $[6.50,\,7.84]\times 10^{-4}$ \\
375  & 750  & 368.927 & $1.018\times 10^{-3}$ & $[9.20,\,11.17]\times 10^{-4}$ \\
187  & 375  & 368.772 & $1.409\times 10^{-3}$ & $[1.273,\,1.546]\times 10^{-3}$ \\
93   & 187  & 368.576 & $2.082\times 10^{-3}$ & $[1.885,\,2.279]\times 10^{-3}$ \\
46   & 93   & 367.986 & $3.504\times 10^{-3}$ & $[3.177,\,3.831]\times 10^{-3}$ \\
\hline
\end{tabular}
\caption{Validation results for the data-augmentation likelihood estimator. The table reports the estimated log-likelihood, the relative error over the log-likelihood, and the corresponding 95\% confidence interval for the relative error. The confidence intervals are constructed by repeating the simulation $250$ times.}
\label{tab:da_validation}
\end{table}

\section{Three-factor short rate model} \label{sect:ngnf_inf_BDFS_subsection}

We further validate the scalability of our surrogate model by considering the time-homogeneous trivariate BDFS model, as introduced in \cite{BDFS}. The BDFS is an extension of the univariate short rate model, incorporating both a stochastic long-run mean and a stochastic volatility. The long-run mean is modeled by an Ornstein-Uhlenbeck (OU) process, while a CIR process describes the volatility.
This model is irreducible and reads
\begin{equation} \label{BDFS_SDE}
    \begin{cases}
        dV(t) = k_1(\alpha_1-V(t))dt + \sigma_1\sqrt{V(t)}dW_1(t) \\
        dO(t) = k_2(\alpha_2-O(t))dt +\sigma_2dW_2(t) \\
         dY(t) = k_3(O(t)-Y(t))dt + \sqrt{V(t)}(\rho dW_1(t)+\sqrt{1-\rho^2}dW_3(t)) \\
        V(0)=v_0,\:O(0)=o_0,\:Y(0)=y_0,
    \end{cases}
\end{equation}
where $W_1(t)$, $W_2(t)$ and $W_3(t)$ are three independent Brownian motions.

We simulate a trajectory of length $n+1=350$ with $\Delta=0.5$ from \eqref{BDFS_SDE} using the same strategy as for \eqref{eq:ngnf_inf_Heston_model}: we introduce 5000 sub-intervals per sampling interval and discard 4999 of them. We generate 700 observations and drop the first 350. We use the following parameters $k_1=3,\:\alpha_1=0.1,\:\sigma_1=0.25,\:k_2=7,\:\alpha_2=0.06,\:\sigma_2=0.03,\:k_3=10,\:\rho=0.5$, which we denote by $\boldsymbol{\mu}^{\star}\in\mathbb{R}^{8}$. Thus, the residual in equation \eqref{eq:ngnf_inf_J_cost_functional_Neural_Galerkin} is minimized over an $8+3+3=14$ dimensional space.

Denoting with $\boldsymbol{x}=[v,o,y]$, the Fokker Planck equation associated with \eqref{BDFS_SDE} is
\begin{equation} \label{FP_BDFS}
    \begin{array}{ll}
    \begin{cases}
        \partial_t{\rho(\boldsymbol{x}|t)} + k_3(o - y)\partial_y{\rho(\boldsymbol{x}|t)} \\
        + k_2(\alpha_2 - o)\partial_o{\rho(\boldsymbol{x}|t)} 
        + k_1(\alpha_1-v)\partial_v\rho(\boldsymbol{x}|t) \\
        -(k_1+k_2+k_3)\rho(\boldsymbol{x}|t) \\
        -\frac{1}{2}\big[ v\partial_{y}^{2}\rho(\boldsymbol{x}|t)
        + \sigma_2^2 \partial_{o}^{2}\rho(\boldsymbol{x}|t) 
        + \sigma_1^2 v\partial_{v}^{2}\rho(\boldsymbol{x}|t) \\
        + 2\rho\sigma_1 v\partial_{yv}^{2}\rho(\boldsymbol{x}|t) \big] 
        -(\rho\sigma_1 \partial_{y}\rho(\boldsymbol{x}|t)
        \\ + \sigma_1^{2}\partial_{v}\rho(\boldsymbol{x}|t)) = 0 
        & \text{for $(v,o,y)\in\mathbb{R}^+\times\mathbb{R}\times\mathbb{R},\:t\in(0,T]$}, \\[6pt]
        \rho(v=0,o,y|t,0,\boldsymbol{x_0},\boldsymbol{\mu}) = 0 
        & \text{for $(o,y)\in\mathbb{R}\times\mathbb{R},\:t\in(0,T]$}, \\
        \rho(\boldsymbol{x}|0,0,\boldsymbol{x_0},\boldsymbol{\mu}) = \delta_{x_0}(\boldsymbol{x}) 
        & \text{for $(v,o,y)\in\mathbb{R}^+\times\mathbb{R}\times\mathbb{R},\:t\in\{0\}$}.
    \end{cases}
    \end{array}
\end{equation}
We utilize a flat prior restricted to the parameters' domain
\begin{equation} \label{prior_bdfs}
    {\pi}_0(\boldsymbol{\mu})=
    \mathrm{1}_{\{\alpha_1>0\}}
    \mathrm{1}_{\{k_1>0\}}
    \mathrm{1}_{\{\sigma_1\in(0,\sqrt{2\alpha_1k_1})\}}
    \mathrm{1}_{\{\alpha_2>0\}}
    \mathrm{1}_{\{k_2>0\}}
    \mathrm{1}_{\{\sigma_2>0\}}
    \mathrm{1}_{\{k_3>0\}}
    \mathrm{1}_{\{-1\leq\rho\leq 1\}},
\end{equation}
and, for the training phase, we sample the parameters $\boldsymbol{\mu}$ by implicitly specifying $\tilde{\pi}_0$ as 
\begin{equation}\label{eq:ngnf_inf_tilde_pi_bdfs}
    \begin{cases}
        k_1 \sim \mathcal{U}(\cdot;[1.5, 4]),\\
        \alpha_1 \sim \mathcal{U}(\cdot;[0.005, 0.25]),\\
        \sigma_1 \sim \mathcal{U}(\cdot;[0.05, 0.4]),\\
        k_2 \sim \mathcal{U}(\cdot;[4, 10]),\\
        \alpha_2 \sim \mathcal{U}(\cdot;[-1.5, 0.25]),\\
        \sigma_2 \sim \mathcal{U}(\cdot;[10^{-3}, 0.25]),\\
        k_3 \sim \mathcal{U}(\cdot;[7.5, 13]),\\
        \rho \sim \mathcal{U}(\cdot;[0.1, 0.9]),
    \end{cases}
\end{equation}
where $\mathcal{U}(\cdot;[a,b])$ denotes the uniform distribution over the interval $[a,b]$.
Samples from $\tilde{\pi}_0$ are further restricted to the prior's support. 

We set the distribution of the initial condition as 
\begin{equation}\label{eq:ngnf_inf_bdfs_initial_distribution}
\eta(\boldsymbol{x}_0)
=
\mathcal{U}(v_0;[0,0.75])\mathcal{U}(o_0;[-2.5,2.5])\mathcal{U}(y_0;[-2.5,2.5]).
\end{equation}
where $\boldsymbol{x}_0=[v_0,o_0,y_0]$.

Like the decomposition of the TPDF of the Heston model \eqref{Decompose_Heston}, we can also decompose the TPDF of the BDFS model as 
\begin{equation} \label{Decompose_BDFS}
    p_{V,O,Y}(v,o,y|v_0,o_0,y_0) = p_{V}(v|v_0,o_0,y_0)p_{O}(o|v_0,o_0,y_0)p_{Y|V,O}(y|v,o,v_0,o_0,y_0).
\end{equation}

The transport map $\boldsymbol{n}_{\theta}(v,o,y)$ is constructed so that the volatility $v$ and the long term-mean $o$ are updated independently from all the other variables. At the same time, the asset price $y$ is modified by conditioning on $v$ and $o$ (which implies setting $m=2$ in the architecture of the Normalizing Flow). 

The reference distribution is picked as a uniform distribution over $[0, 1]$ for the components of
the transport map related to $o$ and $y$, while a Gamma distribution truncated over $[0, 1]$ serves as the
reference density for the first component of $\boldsymbol{n}_{\theta}$.
The linear scaling defined in equation \eqref{eq:ngnf_inf_scaling_last_layer} is used to map the variables into the range \([0,1]^d\) before applying the reference distributions.

The computational parameters and training details are summarized in Table \ref{tab:bdfs_nn_parameters}. 
\begin{table}[!htb]
\centering
\begin{tabular}{lll}
\hline
\textbf{Quantity} & \textbf{Symbol} & \textbf{Value} \\
\hline
Diffusion domain & $\Omega$ & $\mathbb{R}^+\times\mathbb{R}\times\mathbb{R}$ \\
Parameter of the support of the NF (Volatility) & $[\hat{a}_1,\hat{b}_1]$ & [0,2] \\
Parameter of the support of the NF (OU process) & $[\hat{a}_2,\hat{b}_2]$ & [-3,3] \\
Parameter of the support of the NF (log-Asset) & $[\hat{a}_3,\hat{b}_3]$ & [-3,3] \\
Initial condition distribution & $\eta(\boldsymbol{x}_0)$ & See \eqref{eq:ngnf_inf_bdfs_initial_distribution} \\
Parameter distribution & $\tilde{\pi}_0(\boldsymbol{\mu})$ & See \eqref{eq:ngnf_inf_tilde_pi_bdfs} \\
Distribution to approximate $J$ \eqref{eq:ngnf_inf_J_cost_functional_Neural_Galerkin} & $\nu_{{\theta}}(\cdot|\cdot,\cdot)$ & $\Rho(\cdot|\boldsymbol{\theta},\cdot,\cdot)$ \\
Number of layers & $L$ & $5$ (Process $Y$) and $4$ ($O,\:V)$ \\
MTG elements per layer & -- & $5$ for $l>2$ \\
Number of GRU layers & -- & $4$ \\
Hidden size of each GRU layer & -- & $8$ (Process $Y$) and $6$ ($O,\:V$) \\
Dimension of $\boldsymbol{\theta}$ & $M$ & $14448$ \\
Number of Monte Carlo samples to approximate \eqref{eq:ngnf_inf_J_cost_functional_Neural_Galerkin} & $-$ & $17338$ \\
Integrator of the dynamics $\boldsymbol{\dot{\theta}}$ \eqref{eq:ngnf_inf_optimization_Neural_Galerkin} & -- & RK 5(4) \\
Offline GPU Training time & -- & $\simeq 54\mathrm{h}$ \\
Burn-in length & -- & 3000 \\
Length of the Markov chain & -- & 10000 \\
\hline
\hline
\end{tabular}
\caption{
Computational setup, Normalizing Flow architecture, and parameter-sampling details for the BDFS model \eqref{BDFS_SDE}.}
\label{tab:bdfs_nn_parameters}
\end{table}

\subsubsection{Log-likelihood and posterior analysis} \label{Posterior_BDFS}

\begin{figure}[!htb]
\centering

\begin{minipage}[t]{0.27\textwidth}
    \centering
    \includegraphics[width=\linewidth]{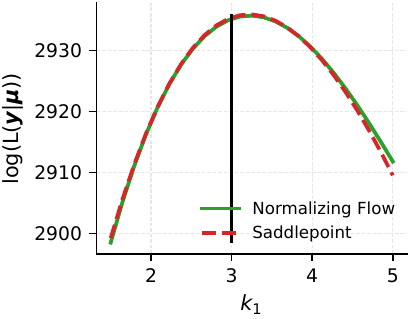}
\end{minipage}\hfill
\begin{minipage}[t]{0.27\textwidth}
    \centering
    \includegraphics[width=\linewidth]{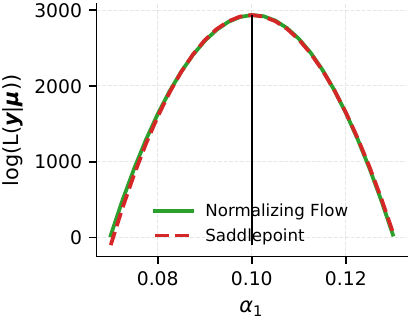}
\end{minipage}\hfill
\begin{minipage}[t]{0.27\textwidth}
    \centering
    \includegraphics[width=\linewidth]{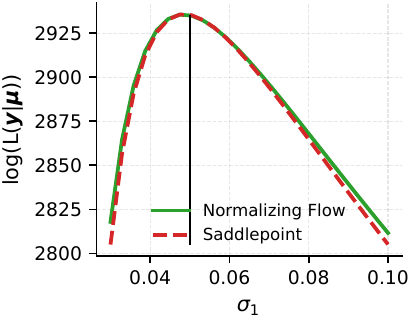}
\end{minipage}

\vspace{0.8em}

\begin{minipage}[t]{0.27\textwidth}
    \centering
    \includegraphics[width=\linewidth]{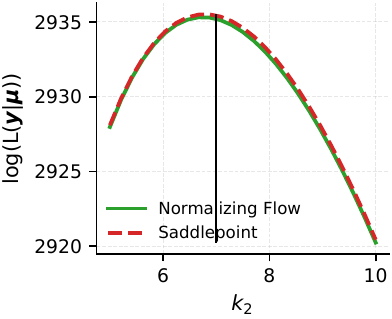}
\end{minipage}\hfill
\begin{minipage}[t]{0.27\textwidth}
    \centering
    \includegraphics[width=\linewidth]{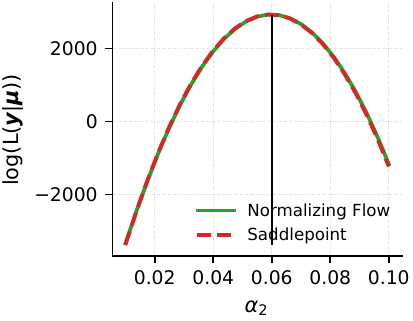}
\end{minipage}\hfill
\begin{minipage}[t]{0.27\textwidth}
    \centering
    \includegraphics[width=\linewidth]{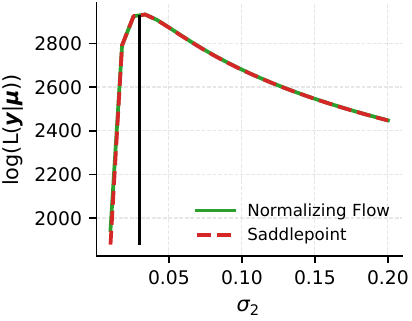}
\end{minipage}

\vspace{0.8em}

\makebox[\textwidth][c]{%
    \begin{minipage}[t]{0.27\textwidth}
        \centering
        \includegraphics[width=\linewidth]{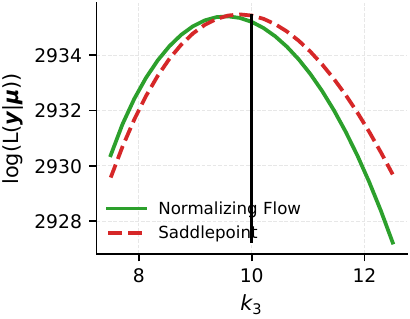}
    \end{minipage}\hspace{0.04\textwidth}
    \begin{minipage}[t]{0.27\textwidth}
        \centering
        \includegraphics[width=\linewidth]{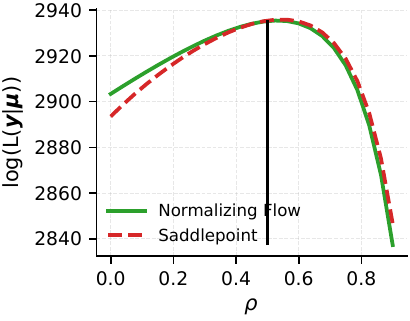}
    \end{minipage}
}

\caption{Log-likelihood evaluated on data generated from the BDFS model \eqref{BDFS_SDE} with parameter $\boldsymbol{\mu}^{\star}$, varying one component of $\boldsymbol{\mu}^{\star}$ at a time.
The different plots show the dependence on each parameter when the others are fixed to the true value of $\boldsymbol{\mu}^{\star}$.
The log-likelihood computed via saddlepoint approximation is depicted with red dashed lines, while the log-likelihood approximated by the Normalizing Flow is shown in green. A vertical line centered on the true parameter is shown in black.}
\label{fig:ngnf_inference_NG_vs_Fourier_log_lh_bdfs}
\end{figure}

In Figure \ref{fig:ngnf_inference_NG_vs_Fourier_log_lh_bdfs}, we show how the log-likelihood varies with respect to each parameter (having fixed all the others to the true values of $\boldsymbol{\mu}^{\star}$).
We compare our Normalizing Flow against a saddlepoint approximation derived from the explicitly available characteristic function (see, e.g., \cite{Daniels1954}). Our results show good agreement, suggesting that our Normalizing Flow model is a promising surrogate, even in high-dimensional settings.

\begin{figure}[!htb]
\centering

\begin{minipage}[t]{0.27\textwidth}
    \centering
    \includegraphics[width=\linewidth]{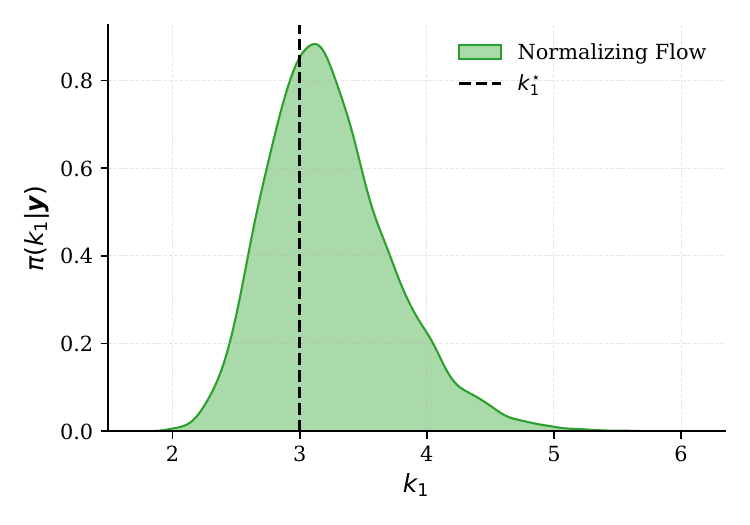}
\end{minipage}\hfill
\begin{minipage}[t]{0.27\textwidth}
    \centering
    \includegraphics[width=\linewidth]{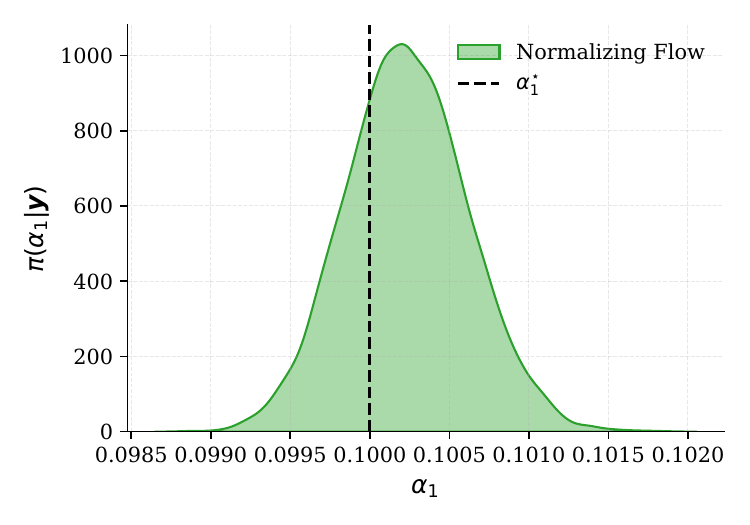}
\end{minipage}\hfill
\begin{minipage}[t]{0.27\textwidth}
    \centering
    \includegraphics[width=\linewidth]{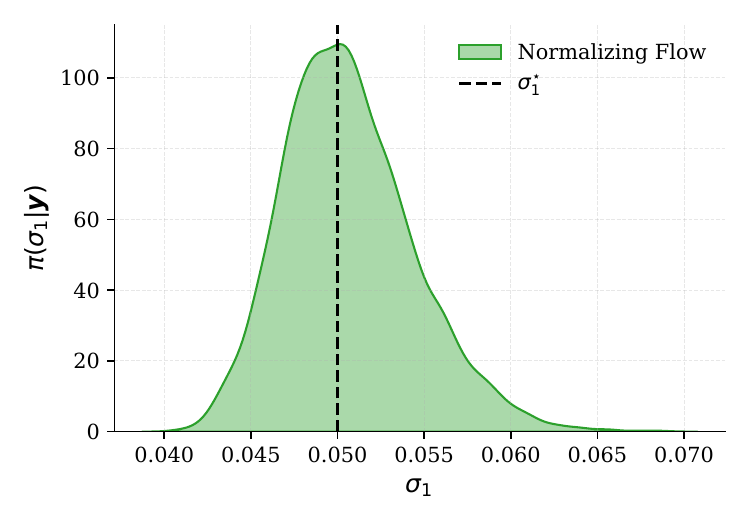}
\end{minipage}

\vspace{0.8em}

\begin{minipage}[t]{0.27\textwidth}
    \centering
    \includegraphics[width=\linewidth]{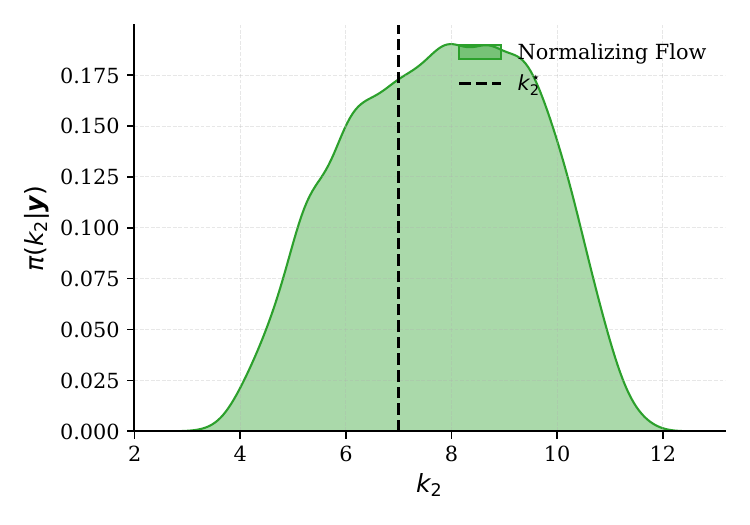}
\end{minipage}\hfill
\begin{minipage}[t]{0.27\textwidth}
    \centering
    \includegraphics[width=\linewidth]{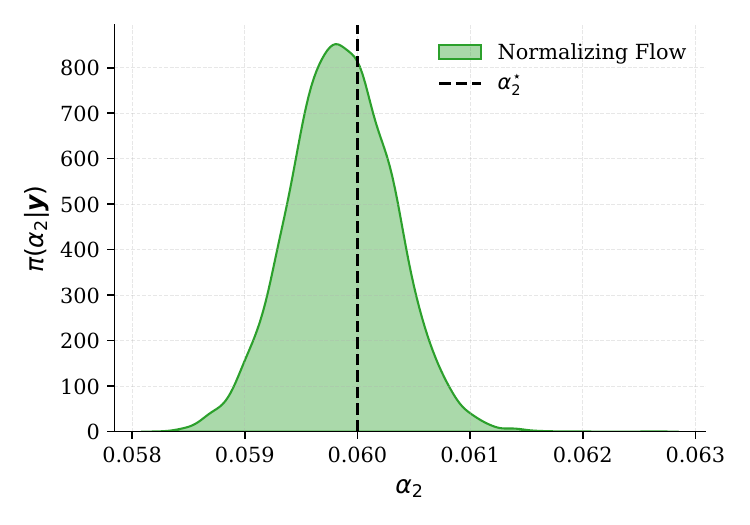}
\end{minipage}\hfill
\begin{minipage}[t]{0.27\textwidth}
    \centering
    \includegraphics[width=\linewidth]{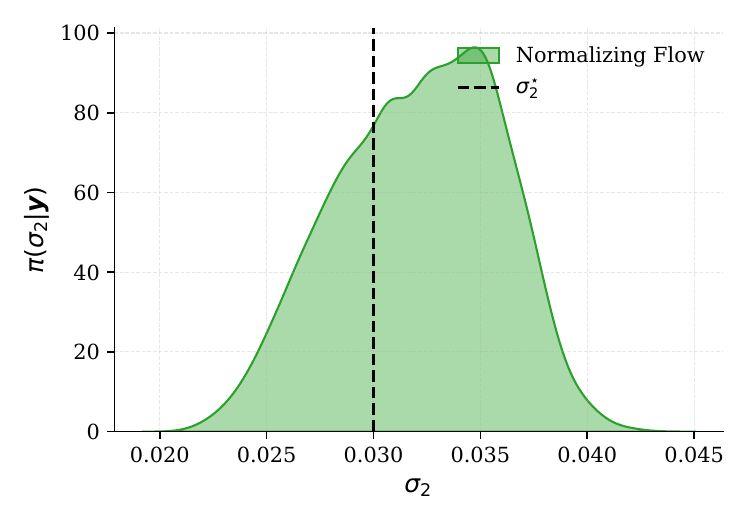}
\end{minipage}

\vspace{0.8em}

\makebox[\textwidth][c]{%
    \begin{minipage}[t]{0.27\textwidth}
        \centering
        \includegraphics[width=\linewidth]{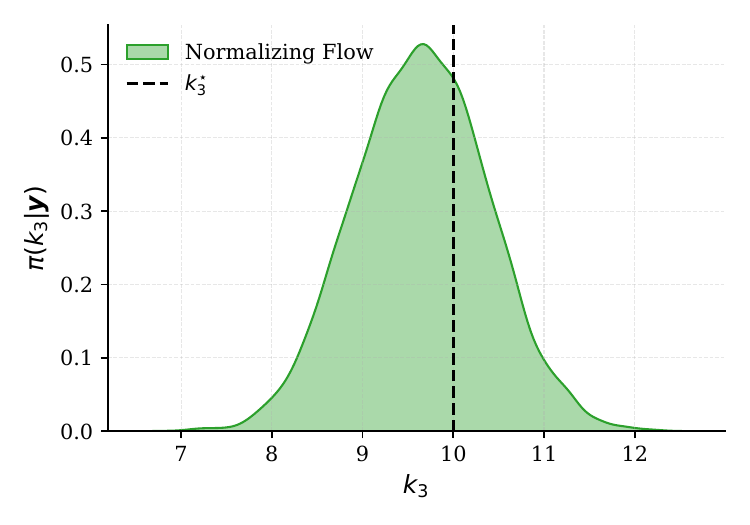}
    \end{minipage}\hspace{0.04\textwidth}
    \begin{minipage}[t]{0.27\textwidth}
        \centering
        \includegraphics[width=\linewidth]{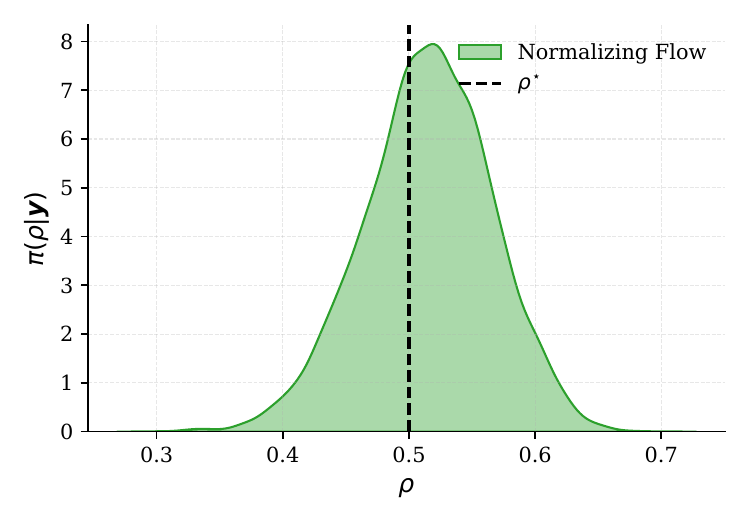}
    \end{minipage}
}

\caption{Posterior marginal distributions for the parameters of the BDFS model \eqref{BDFS_SDE}. The chain is composed of 10000 samples.}
\label{fig:ngnf_inference_posterior_bdfs_marginals}
\end{figure}

Finally, we draw samples from the posterior distribution using MCMC in the form of Slice Sampling with axis-aligned. For this example, we do not compare the NF's performance to any reference, as targeting the posterior via the saddlepoint approximation proved computationally intractable. 
The marginal posterior distributions, presented in Figure \ref{fig:ngnf_inference_posterior_bdfs_marginals}, assign meaningful uncertainty to the model's parameters, indicating that the Normalizing Flow appears scalable to higher-dimensional benchmarks.

\section{Sampling from the Truncated-bounded Normalizing Flow}

    In general, when there are multiple elements in the mixture of Gaussians \eqref{eq:ngnf_inf_base_transformation}, we lose the analytic invertibility of $\boldsymbol{n}_{\theta}$, and sampling from \eqref{eq:ngnf_inf_FP_approximation_by_NF} requires the solution of nonlinear equations. We consider two alternatives for sampling from the Normalizing Flow: a high-fidelity root-finding method and an approximate method that is computationally cheaper.
    Here, we approximate the target density $\Rho$ by replacing the bijection of layer $l$, for $l\geq2$, with the CDF of a unique element of the mixture of truncated Gaussian random variables. This element is chosen with probability given by its mixture weight. The entire procedure is reported in Algorithm \ref{approximated_sample_from_NF}.
    We remark that the parameters $\boldsymbol{x_0}$ and $\boldsymbol{\mu}$ influence the sampling step, as they define the mean, variances, and mixture weight characterizing the mixture of Gaussians of each layer.
   \begin{algorithm}[t]
    \caption{Sampling from $\Rho(\cdot |\boldsymbol{\theta}(\tau), \boldsymbol{x}_0, \boldsymbol{\mu})$}
    \label{approximated_sample_from_NF}
    \begin{algorithmic}[1]
    \Require $\tau$: relative time instant of the Normalizing Flow; $\boldsymbol{x}_0$; $\boldsymbol{\mu}$
    \Ensure Sample $\boldsymbol{x}\sim \Rho(\cdot\mid\boldsymbol{\theta}(\tau),\boldsymbol{x_0},\boldsymbol{\mu})$
    \State $\boldsymbol{z}^{L} \sim \boldsymbol{Z}$ \Comment{Draw a sample from the reference distribution}
    
    \For{$m = 0, \dots, d-1$}
        \For{$l = L, \dots, 2$}
            \State $\tilde z_{m+1}^{\,l} = \dfrac{z_{m+1}^{\,l} + 1}{2}$
            \State $[\boldsymbol{\alpha}^{l,m+1}(\tau), \boldsymbol{\mathrm{m}}^{l,m+1}(\tau), \boldsymbol{\sigma}^{l,m+1}(\tau)] = \mathrm{NN}^{l,m+1}(\boldsymbol{x}_{1:m}, \boldsymbol{x}_0, \boldsymbol{\mu})$
            \Comment{Compute the parameters of the MTG}
    
            \If{Exact Sampling}
                \State Solve for $z_{m+1}^{\,l-1}$ in
                \[
                \tilde z_{m+1}^{\,l}
                =
                \sum_{k=1}^{G_{l,m+1}}
                \alpha_k^{l,m+1}(\tau)\,
                \Phi\left(
                z_{m+1}^{\,l-1}
                \mid
                \mathrm{m}_k^{l,m+1}(\tau),
                \sigma_k^{l,m+1}(\tau),
                -1,1
                \right)
                \]
                using a root-finding algorithm
            \Else
                \State $k \sim \mathcal{A}(\cdot |\boldsymbol{\alpha}^{l,m+1})$
                \Comment{Select one index from the discrete distribution on $\{1,\dots,G_{l,m+1}\}$}
                \State $z_{m+1}^{\,l-1} =
                \Phi^{-1}\left(
                \tilde z_{m+1}^{\,l}
                \mid
                \mathrm{m}_k^{l,m+1},
                \sigma_k^{l,m+1},
                -1,1
                \right)$
                \Comment{Analytical inversion of the selected truncated Gaussian CDF}
            \EndIf
        \EndFor

        \State $x_{m+1} = \dfrac{z_{m+1}^{\,1}+1}{2}\,(\hat{b}_{m+1}-\hat{a}_{m+1}) + \hat{a}_{m+1}$
        \Comment{Rescale the sample to the NF's support $[\hat{a}_{m+1}, \hat{b}_{m+1}]$}
    \EndFor
    
    \State \Return $\boldsymbol{x}$
    \end{algorithmic}
    \end{algorithm}

     The approximate method is not designed to replicate the exact sampler with arbitrary precision, but rather to generate representative training points at a much lower computational cost.
     Since it is solely employed to generate samples for evaluating the training residual, it does not need to be exact. 
    During testing, to obtain more reliable samples from the model, root-finding routines can be implemented instead.

\subsection{Approximated sampling from Normalizing Flows: validation}\label{Approximated_samples_NF_appendix}

We will empirically evaluate whether the approximate samples generated by Algorithm \ref{approximated_sample_from_NF} are sufficiently representative to serve as training points for estimating the expected residual. Our validation procedure consists of two main steps:

\begin{itemize}
    \item After the training phase, given a prescribed initial condition $\boldsymbol{x_0}\sim\eta$, parameter vector $\boldsymbol{\mu}\sim\tilde{\pi}_0$, and time horizon $\Delta$, we use Algorithm \ref{approximated_sample_from_NF} to obtain $N$ samples from the Normalizing Flow.  

     \item We push these samples back through the Normalizing Flow using the map $\boldsymbol{n}_{\theta^{\star}}$ and check whether their distribution aligns with the Normalizing Flow's reference distribution.
\end{itemize}

To maintain a moderately high-dimensional framework, we conduct this validation phase on the Normalizing Flow trained on the model described in section \ref{sect:ngnf_inf_BDFS_subsection}. This test assesses whether the map correctly transforms the samples into the intended reference distribution of the Normalizing Flow. We consider the third component of the transport map, which approximates the conditional distribution of the log-asset price, namely a uniform distribution over $[0,1]$. 

We draw 400 samples from the Normalizing Flow conditioned on the following parameters: $\boldsymbol{\mu}=\boldsymbol{\mu}^{\star}$ and $\boldsymbol{x_0}=[0.0948,0.0559,-0.013]$. We consider a time horizon of $\Delta =0.05$, which is a crucial moment for the optimization routine because the intensity of the singular initial condition is still very concentrated. In the early stages of the optimization process, it is essential to obtain sufficiently representative samples to evaluate the expected residual stably.

\begin{figure}[!htb]
\minipage{0.43\textwidth}
  \includegraphics[width=\linewidth]{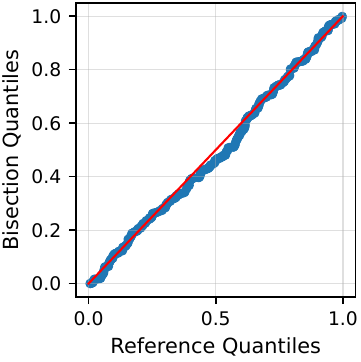}
\endminipage\hfill
\minipage{0.43\textwidth}
  \includegraphics[width=\linewidth]{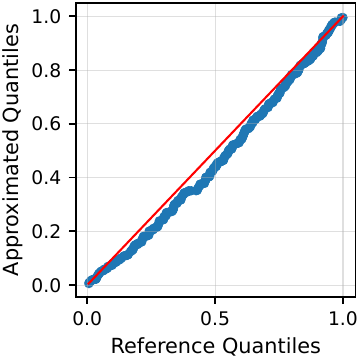}
\endminipage\hfill
\caption{Q-Q (quantile-quantile) plot comparing the reference distribution of the conditional density of the log-asset price approximated with the Bisection-Method (left) and with Algorithm \ref{approximated_sample_from_NF} (right). The Normalizing Flow is parametrized by the weights and biases saved at time 0.05 of Algorithm \ref{alg:ngnf_inf_integrate_NF_map}.}
\label{fig:ngnf_inference_qq_plot_asset_005}
\end{figure}

\begin{figure}[!htb]
\minipage{0.43\textwidth}
  \includegraphics[width=\linewidth]{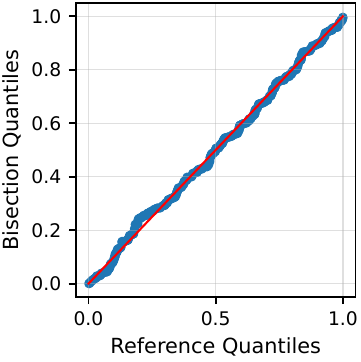}
\endminipage\hfill
\minipage{0.43\textwidth}
  \includegraphics[width=\linewidth]{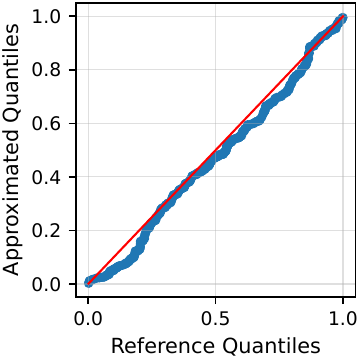}
\endminipage\hfill
\caption{Q-Q (quantile-quantile) plot comparing the reference distribution of the conditional density of the log-asset price approximated with the Bisection-Method (left) and with Algorithm \ref{approximated_sample_from_NF} (right). The Normalizing Flow is parametrized by the weights and biases saved at time 0.5 of Algorithm \ref{alg:ngnf_inf_integrate_NF_map}.}
\label{fig:ngnf_inference_qq_plot_asset_t_05}
\end{figure}

We conduct the same test using a high-fidelity Bisection method, which effectively approximates the inverse of the transport map $\boldsymbol{n}_{\theta}$ within a specified tolerance. 
In Figure \ref{fig:ngnf_inference_qq_plot_asset_005}, we compare the quantiles obtained by the two methods. We also repeat the test at a later time, $\Delta=0.5$, which occurs at the conclusion of the optimization routine. The results for this test are illustrated in Figure \ref{fig:ngnf_inference_qq_plot_asset_t_05}.

The quantiles reconstructed using our algorithm exhibit greater deviation from the reference distribution than those obtained with the Bisection method. However, the computational effort required to run Algorithm \ref{approximated_sample_from_NF} is significantly lower, especially when aiming for very small tolerances with the root-finding algorithm.
As previously mentioned, our algorithm does not need to be exact within a specific tolerance; instead, it aims to provide samples that represent the general behavior of the reference distribution. As illustrated in Figures \ref{fig:ngnf_inference_qq_plot_asset_005} and \ref{fig:ngnf_inference_qq_plot_asset_t_05}, the quantiles derived from our approximated algorithm align reasonably well with the true quantiles. Although we do not investigate this point further here, using samples close to the support of the true distribution, but not necessarily sampled from it, may be beneficial for estimating the expected residual. Indeed, they may act as contrastive samples, preventing the numerical solution from collapsing onto the training points. This collapse could occur because the null solution minimizes the residual linked to the Fokker-Planck equation. 
    This sampling strategy can be further refined using speculative sampling techniques \cite{leviathan2023fast}.
By selecting a simplified target architecture with only one Gaussian per layer as a draft model, we can accept or reject its proposals to draw exact samples from the Normalizing Flow at a significantly lower cost.

\end{document}